\documentclass[preprint,12pt,authoryear]{elsarticle}
\usepackage{times,latexsym}
\usepackage{multirow}
\usepackage{tikz}
\usepackage{setspace}
\usepackage{array}
\usepackage{lipsum}
\usepackage{amsmath}
\usepackage{amsfonts}
\usepackage{pifont}
\usepackage{booktabs}
\usepackage{url}
\usepackage{graphicx}
\usepackage{caption}
\usepackage{subcaption}
\usepackage[T1]{fontenc}
\newcommand{\quotes}[1]{``#1''}
\newtheorem{definition}{Definition}

\usepackage{amssymb}

\journal{Computer Speech \& Language}

\begin{document}

\begin{frontmatter}



\title{End-to-End Speech-to-Text Translation: A Survey}

\author[]{Nivedita Sethiya, Chandresh Kumar Maurya}
\affiliation[label1]{organization={Indian Institute of Technology Indore},
            country={India}}


\author{}


\begin{abstract}
Speech-to-Text (ST) translation pertains to the task of converting speech signals in one language to text in another language. It finds its application in various domains, such as hands-free communication, dictation, video lecture transcription, and translation, to name a few. Automatic Speech Recognition (ASR), as well as Machine Translation(MT) models, play crucial roles in traditional ST translation, enabling the conversion of spoken language in its original form to written text and facilitating seamless cross-lingual communication. ASR recognizes spoken words, while MT translates the transcribed text into the target language. Such integrated models suffer from cascaded error propagation and high resource and training costs. As a result, researchers have been exploring end-to-end (E2E) models for ST translation. However, to our knowledge, there is no comprehensive review of existing works on E2E ST. The present survey, therefore, discusses the works in this direction. We have attempted to provide a comprehensive review of models employed, metrics, and datasets used for ST tasks,  providing challenges and future research direction with new insights. We believe this review will be helpful to researchers working on various applications of ST models.
\end{abstract}




\end{frontmatter}


\section{Introduction}
The Speech-to-Text (ST) translation task aims to convert a speech in one language into text in another language. It finds its applications in various areas such as \emph{automatic subtitling, dictations, video lecture translations, tourism, telephone conversations}, to name a few. There are many facets under which the ST problem can be cast. For example, are we performing ST translation online (aka simultaneous translation) or offline? The former is required in live video streaming, while the latter is helpful for movies where some latency may be allowed. The ST problem is further exacerbated by noisy inputs, low-resource/code-mix languages, and the presence of multiple speakers.

\begin{figure}
    \centering
    \includegraphics[width=\textwidth]{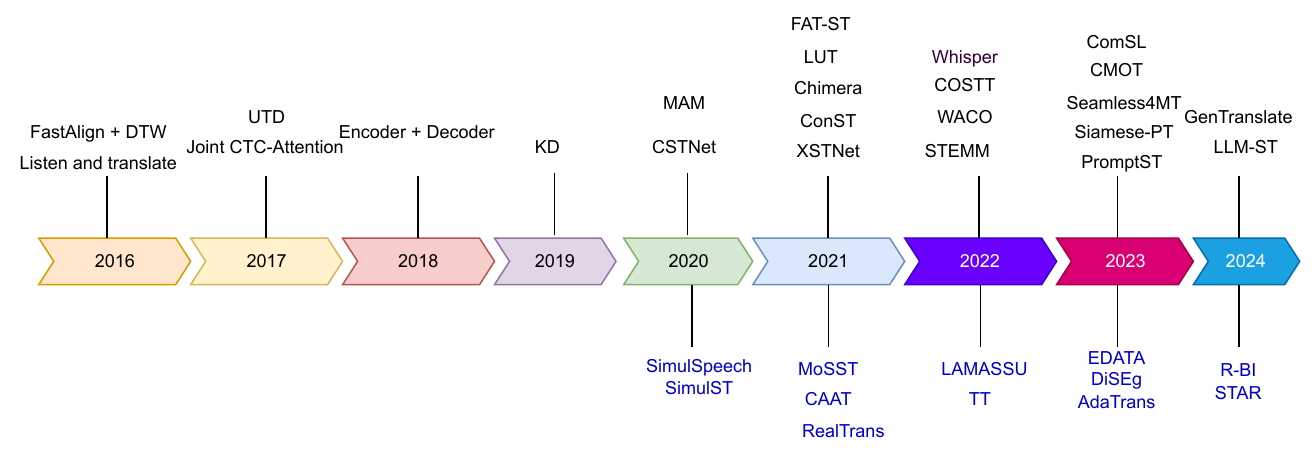}
    \caption{History of E2E ST models. Blue color models correspond to streaming models discussed in \S \ref{stinstreaming}. Note that here we have listed only a few selected representative models.}
    \label{fig:histroy}
\end{figure}
Historically, the ST problem has been solved by pipelining ASR and MT models together where ASR models take speech in a source language as input and generate the transcript. Whereas MT models translate the transcript into the target language. Such a cascade model suffers from problems like  \emph{error propagation, higher training, and inference latency}. Therefore, the current trend in developing the ST model is toward the E2E system which is defined as
\begin{definition}
     A unified E2E ST model is implemented, facilitating combined training and recognition processes aimed at consistently reducing the anticipated error rate, thereby bypassing the need for independently acquired sources of knowledge.
\end{definition}
Therefore, the main goal of the E2E ST model is to achieve a reduced error rate, with secondary objectives potentially including decreased training/inference duration and memory usage. 

There has been a lot of work building E2E ST models (as shown in fig. \ref{fig:histroy}), datasets, and metrics in recent years. However, a systematic and comprehensive review of E2E ST works is missing. The authors found that a review paper \citep{Xu2023RecentAI} on ST was published recently. The review mentioned above categorizes existing works mainly based on modeling, data, and application issues. They do not cover the data sets available for the ST tasks nor provide any insights into the cascade vs. E2E model performances. Also, the future open problems provided by them are limited. On the other hand, our work comprehensively reviews the existing models for ST tasks, evaluation methods, metrics, and datasets from a completely different perspective and critically analyzes the existing works; after that, we identify several challenges and future research directions. Thus, our work may be deemed complimentary to \citep{Xu2023RecentAI}. 

\begin{figure}[t]
    \centering
    \includegraphics[width=\textwidth,height=10cm]{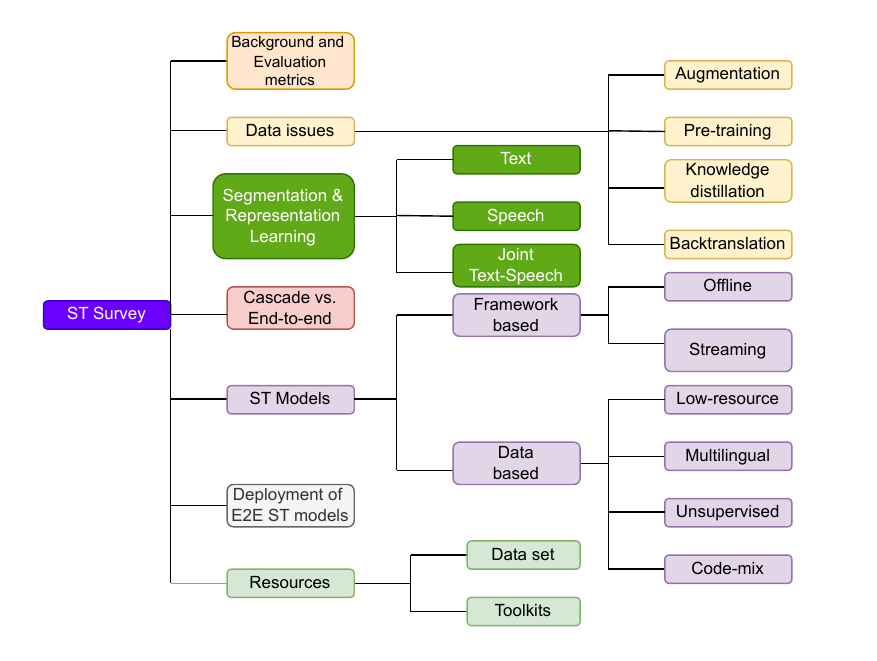}
    \caption{Organization of the survey paper}
    \label{fig:organization}
\end{figure}
The following review is structured following the taxonomy in fig. \ref{fig:organization}.  In \S \ref{bck}, we establish the foundation of the ST task through a formal definition, and we subsequently delve into the various metrics and loss functions adopted by different researchers in \S \ref{evaluation}. A comparative discussion between cascade and end-to-end models is presented in \S \ref{casvse2e}. Training of E2E ST models suffers from data issues and how to combat them is elaborated in \S \ref{dataissues}. Speech and Text segmentation and representation is an important task in ST model development discussed in \S \ref{segrepr}.  In \S \ref{endtoend}, we delve into the strategies employed to tackle the ST problem. We categorize these approaches based on the frameworks utilized and the characteristics of the data involved. Data and toolkits required for ST modeling are discussed in \S \ref{resources}. Finally, in \S \ref{future}, we explore the prospects for future research and open problems within the field. 


\section{Background} \label{bck}
This section describes the ST task formally and presents the loss functions and evaluation metrics commonly employed to optimize ST models.

\subsection{Task Definition}
ST task can be defined as translating the given input speech $U$ in one language to translated text $V$ in another language with the transcription text $X$ (optionally). Formally, it is defined as follows: Given a dataset $D=\{({\bf u}^i,{\bf x}^i, {\bf v}^i)|i=1,2,\ldots, n\}$ of pairs of input speech features  ${\bf u}=(u_1,u_2,\ldots, u_{T_u})$ in a language and output text tokens  ${\bf v}=(v_1,v_2,\ldots, v_{T_v})$ in a different language, the objective of the ST task is to minimize the conditional probability given below:
\begin{equation}\label{autoreg}
    p({\bf v}| {\bf u}; \theta) =\prod_{t=1}^{T_v} p(v_t|v_{<t}, {\bf u};\theta) 
\end{equation}
In the above equation, $T_u$, $T_v$, and $\theta$ are the lengths of input features, the number of output tokens, and the model parameter, respectively.
Note that the problem formulation given in \eqref{autoreg} is for \emph{Autoregressive} (AR) models \footnote{Non-autoregressive (NAR) models are an alternative modeling approach that has been proposed in the past few years for the ST task. Only a sparse number of works exist in the literature. We discuss NAR briefly in \S \ref{frameworks}}.
Usually, it is assumed that there are $n$ parallel speech-text pairs in our corpus, and the model is optimized for negative log-likelihood over these pairs as
\begin{equation}
    \ell(\theta|D) = -\sum_{i=1}^n \log P({\bf v}^i| {\bf u}^i; \theta)
\end{equation}
The above optimization is usually solved using an encoder-decoder with an attention approach. Essentially, an encoder maps speech input to a hidden state representation $h$ followed by a decoder which takes the previously generated text tokens $v_{<t}$, encoder hidden state $h$ and attention vector ${\bf \alpha}$ \citep{Vaswani2017AttentionIA}. Offline ST translation can look at the whole speech before producing output text tokens, whereas streaming ST can start translation of partial speech signal. 

\section{Evaluation Metrics}\label{evaluation}
This section discusses various metrics used to evaluate the E2E ST models. The metrics to evaluate E2E ST models are categorized into two types: {\bf quality} and {\bf latency}. The quality of the E2E ST models is the measure of how close the ST translation is to the target sentence. The latency is the time elapsed between the pronunciation of a word and the generation of its textual translation.

\subsection{ Quality-based metrics}
The quality-based metrics measure how close the translation is to the target sentence. Most of the existing literature evaluates these scores on \emph{detokenized} output which is the string formed by combining the tokens. Standard metrics for evaluating ST task performance are commonly used MT evaluation metrics such as  Bi-lingual Evaluation Understudy (BLEU) \citep{papineni2002bleu}, Translation Error Rate (TER) \citep{snover2006study} via sacreBLEU, Metric for Evaluation of Translation with Explicit word Ordering (METEOR) \citep{banerjee2005meteor}, and CHaRacter-level F-score (CHRF), and CHRF++ \citep{popovic2015chrf}. Recently BERTScore has shown promising results on comparing with human evaluations. The BERTScore \citep{zhang2019bertscore} is an automatic evaluation metric that scores the similarity between the translated text and the referenced text. It takes into account the Recall, Precision, and Fscore. There are a few other evaluation metrics such as TRANSTAC \citep{schlenoff2009evaluating} and  which are less frequently reported.

\subsection{ Latency-based metrics}
For streaming ST tasks, researchers report a metric for measuring \emph{latency}, which is defined as the delay incurred in starting to produce the translation.
Let ${\bf u},{\bf v}$ and $\hat{{\bf v}}$ denote the input speech sequence, ground truth text sequence, and system-generated hypothesis sequence, respectively. In the streaming ST task,  models produce output with partial input. Suppose ${\bf u}_{1:t}=\{(u_1,\ldots,u_t), t < T_u\}$ has been read when generating $v_s$, the delay in $v_s$ is defined as \citep{Ma2020SIMULEVALAE}
\begin{equation}
    d_s = \sum_{k=1}^t T_k
\end{equation}
where $T_k$ is the duration of the speech frame $u_k$.
The latency metrics are calculated using a method that analyzes a sequence of time delays $[d_1,\ldots,d_{T_v}]$.
\begin{itemize}
    \item {\bf Average Proportion (AP)} \citep{cho2016can} calculates the mean fraction of the source input that is read during the target prediction generating process.
    \begin{equation}
        AP=\frac{1}{T_v  \sum_{k=1}^{T_u} T_k} \sum_{s=1}^{T_v} d_s
    \end{equation}

    \item {\bf Average Lagging (AL)} measures the distance between the speaker and the user based on the number of words used in the conversation \citep{Ma2018STACLST}.
    \begin{equation}
    AL = \frac{1}{\tau(T_u)} \sum_{s=1}^{\tau(T_u)}d_s-\hat{d_s}   
    \end{equation}
    Where $\tau(T_u)=\min \{s \mid d_s=\sum_{k=1}^{T_u} T_k \}$ and $\hat{d_s}$ are the delays of an ideal policy defined as \citep{Ma2020SIMULEVALAE}
    \begin{equation} \label{al}
        \hat{d_s}= (s-1) \sum_{k=1}^{T_u}\frac{T_k}{T_v}
    \end{equation}
     \item {\bf Differentiable Average Lagging (DAL)}  One issue with AL is that it is not differentiable because of the $\min$ function. To solve this, \citep{Cherry2019ThinkingSA} introduces a minimum delay of $1/\gamma$ after each operation and defines DAL as 
    \begin{equation}
        DAL = \frac{1}{T_v} \sum_{s=1}^{T_v} d_s^{'} -\frac{s-1}{\gamma}
    \end{equation}
    where
  \begin{equation}
    d_s^{'} = \begin{cases}
        d_s, & \ s=0 \\
        \max(d_s, d^{'}_{s-1}+\gamma), & s>0
        \end{cases}
    \end{equation}
and $\gamma=T_v /\sum_{k=1}^{T_u} T_k$
    \item {\bf Length-Adaptive Average Lagging (LAAL)} 
    One issue with AL metric for simultaneous translation is that though it can handle the \emph{under-generation}\footnote{Under/Over-generation problem refers to the length of the generated text compared to the reference translation text.} problem, it is unable to handle \emph{over-generation} and produces biased score. To alleviate this issue, \citep{papi-etal-2022-generation} propose LAAL which modifies \eqref{al} as
    \begin{equation} \label{laal}
        \hat{d_s}= (s-1) \sum_{k=1}^{T_u}\frac{T_k}{\max\{T_v,\hat{T_v}\}}
    \end{equation}
    Essentially, it divides  \eqref{al} by the maximum length of the reference and predicted text. As such, it can handle both over and under-generation problems.

  \item {\bf Average Token Delay (ATD)}  
AL metric does not take into account the length of the partial translation output, i.e., it does not consider the latency caused by longer outputs. To remedy this issue, ATD \citep{Kano2023AverageTD}, defined below, has been proposed recently.  
\begin{equation} \label{atd}
    ATD = \frac{1}{T_v} \sum_{s=1}^{T_v}(T(v_s)-T(u_{a(s)}))
\end{equation}
     where
  \begin{align} \label{atd2}
        a(s) &= \min(s-f(s), d_s) \\
   f(s) &= (s-1) - a(s-1)
  \end{align}
 $T (\cdot)$ in \eqref{atd} represents the ending time
of each input or output token. The token is a sub-segment in speech, a character, or a word in text. $a(s)$ represents the index of the input token corresponding to $v_s$ in the time difference calculation and $a(0) =
0$. $f(s)$ in \eqref{atd2} represents how much longer
the duration of the previous translation prefix is
than that of the previous input prefix.

\end{itemize}

\subsection{Loss Functions}
Let $D= (u, x, v)$ be a tuple where $u, x$, and $v$ are the speech, the transcription text, and the translation text, respectively. The following are the various loss functions that are used to optimize the performance of the E2E ST models:
\begin{itemize}
  \item \textbf{Distillation Loss} \citep{liu2019end} The student model not only matches the ground truth, but also the teacher models's output probabilities, which reduces the variance of the gradients.
    \begin{equation}
        L_{KD} = -\sum_{(x,v)\in D} \sum_{t=1}^{N} \sum_{k=1}^{|V|} S(v_{t}=k| v_{<t}, x) \log T(v_{t}=k| v_{<t}, x)
    \end{equation}
    where $S$ and $T$ denote the output distribution of student and teacher models, respectively.
  \item \textbf{CTC Loss} \citep{Ren2020SimulSpeechES} computes the most likely alignment of output text sequence given input speech sequence by summing over the all possible output sequence paths. 
    \begin{equation}
        L_{CTC}=  - \sum_{(u,x) \in D} \sum_{z \in \phi(x)} \log p(z|u)
    \end{equation}

    \item \textbf{Cross-Modal Adaptation Loss} \citep{liu2020bridging} is defined as the sum of all the Mean Squared Errors of the speech and the transcription texts.
    \begin{equation}
        L_{AD}= \Biggl\{ 
        \begin{matrix}
            \sum_{(u,x) \in D} MSE(\bar{h_{u}}, \bar{h_x}); & & $seq-level$\\
            \sum_{(u,x) \in D} MSE(h_{u}, h_x); & & $word-level$
        \end{matrix}
    \end{equation}
    where $h_{u}$ and $h_{x}$ are the speech and word embeddings, and $\bar{h_{u}}$ and $\bar{h_x}$ are the average speech and word embeddings, respectively. MSE represents the difference between the two embeddings.
    
    \item \textbf{Cross-Entropy Loss} \citep{ye2021end} is the negative likelihood of the data combined over all the subtasks such as ASR, MT, ST and also from external-MT. 
    \begin{equation}
        L_{\theta}= - \sum_{x,v\in D' \cup D_{MT-ext}} \log p(x|v; \theta),
    \end{equation}
    where $D'= D_{ASR} \cup D_{MT} \cup D_{ST}$ is the superset of all the parallel subsets data.
 \item \textbf{Contrastive Loss} \citep{ye2022cross} is computed between the speech and the transcription text bringing them closer, and pushing the unrelated pairs farther.
    \begin{equation}
        L_{CON}= - \sum_{(u,x) \in D} \log\frac{\exp({\cos (\bar{h_{u}}, \bar{h_{x}})}/\kappa)}{\sum_{\forall x_j \notin \bar{h_{x}}}\exp({\cos (\bar{h_{u}}, \bar{h_{x}}(x_j))}/\kappa)}, 
    \end{equation}
    where $cos$ and $\kappa$ denote the cosine similarity and temperature hyperparameter, respectively.
      \item \textbf{ST Loss} \citep{ouyang2022waco} is defined as the negative log-likelihood of the translation text given the source speech as follows
    \begin{equation}
        L_{ST} = - \sum_{(u,v) \in D} \log p(v|u)
    \end{equation}

    \item \textbf{MT Loss} \citep{ouyang2022waco} is defined as the negative log-likelihood of the translation text given the source transcript as follows
    \begin{equation}
        L_{MT} = - \sum_{(x,v) \in D} \log p(v|x)
    \end{equation}

    \item \textbf{ASR Loss} \citep{ouyang2022waco} is defined as the negative log-likelihood of the transcription text given the source speech as follows
    \begin{equation}
        L_{ASR} = - \sum_{(u,x) \in D} \log p(x|u)
    \end{equation}
   
\end{itemize}
\begin{figure} \label{cascadevse2e}
    \centering
    \subfloat[]{ \label{fig:cascade}
    \includegraphics[height= 3cm, width= 6cm]{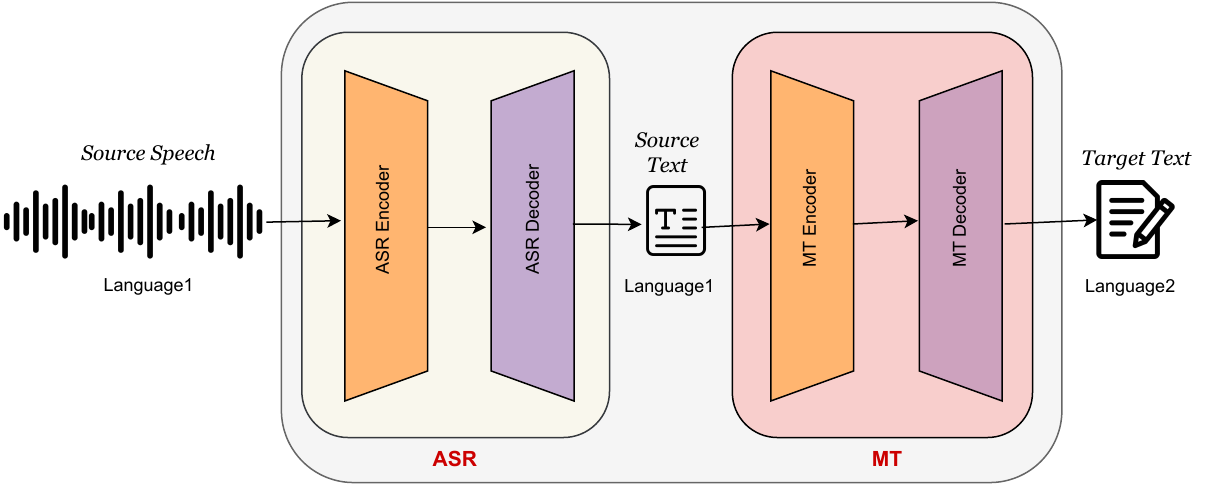}
    }
    \subfloat[]{ \label{fig:e2e}
    \includegraphics[height= 3cm, width=5cm]{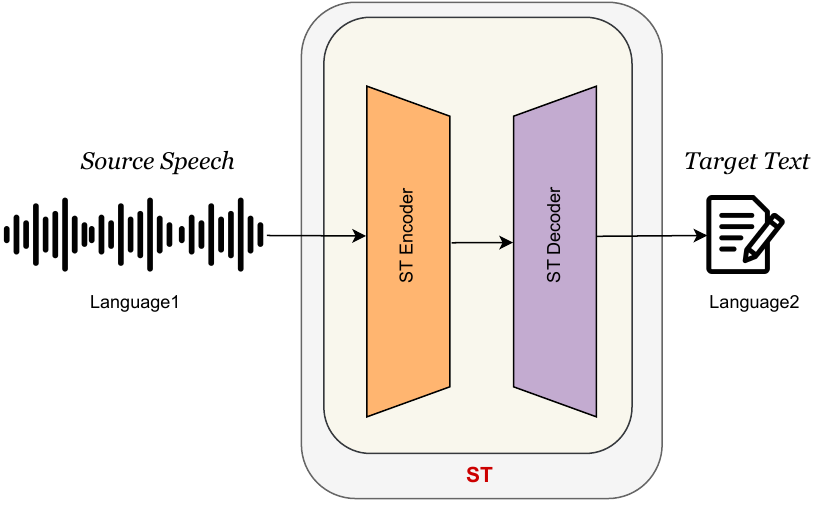}
    }
   
    \caption{ Generic architecture of (a) Cascade  (d) E2E  ST model}
\end{figure}
\section{Cascade vs. End-to-End} \label{casvse2e}
The traditional ST translation methods involve a cascade approach-- First, applying ASR on the given speech and then performing MT on the transcription produced by ASR (see fig. \ref{fig:cascade}) Such a cascade model is prone to several issues, such as (a) error in the ASR model can propagate to the MT model, (b) higher training time, (d) inability to capture non-lexical cues such as prosody,  and (d) resources required for training. To mitigate such issues, various researchers propose using E2E models (see fig. \ref{fig:e2e}) for ST task \citep{berard2016listen, anastasopoulos2016unsupervised,berard2018end,Gangi2019OnetoManyME,Bentivogli2021CascadeVD}. An E2E model offers joint training from scratch; avoids separately trained knowledge sources; and produces the output in a single pass \citep{rohite2easr}. Because of simpler training, lower memory footprint, and cost, E2E model development has gained significant momentum in the research community.

Despite E2E models demonstrating superiority over cascade ST models based on the aforementioned criteria, they still fall short in comparison to the latter in terms of both automatic and human evaluation metrics \citep{etchegoyhen2022cascade, agrawal-etal-2023-findings}. In particular, \citep{Lam2020CascadedMW, etchegoyhen2022cascade} show that the cascade model outperforms E2E in a low-resource setting (Basque $\rightarrow$ Spanish) while employing in-domain and out-of-domain data for training the ASR and MT components. The gap is more significant when models are trained using unrestricted data. However, as shown by \citep{Bentivogli2021CascadeVD} on three language directions, the gap between cascade and E2E is closed, though primarily on English on one side. The same conclusion is found by \citep{Tsiamas2024PushingTL} as well. Another study \citep{Zhou2024ProsodyIC} shows that E2E models can capture para-linguistic features of speech and outperform cascade models in disambiguating wh-phrases. Such a study alludes to further comparative study involving more languages and domains to assert the claim that the performance gap is indeed closed.

\section{Data Issues}\label{dataissues}
The lack of adequate parallel speech-text corpora, essential in large quantities for training direct ST models, significantly impedes the performance of such models. The necessity for supervised ST data poses challenges in applying E2E ST systems to low-resource languages, where creating labeled parallel speech-text corpora demands substantial investments of time, money, and expertise. To address data scarcity, various techniques such as data augmentation, pre-training,  back-translation, knowledge distillation, etc., are employed. These methods are elaborated as follows.
\subsection{Augmentation}
Data augmentation is a technique in machine learning to synthetically create more data points by applying the class-preserving transformations \citep{cui2015data}. The objective is to increase the variability in the data so that the generalization and robustness of the model may be enhanced. Data augmentation can be applied to both speech and text.
\begin{figure*}
\centering
    \includegraphics[height=7cm]{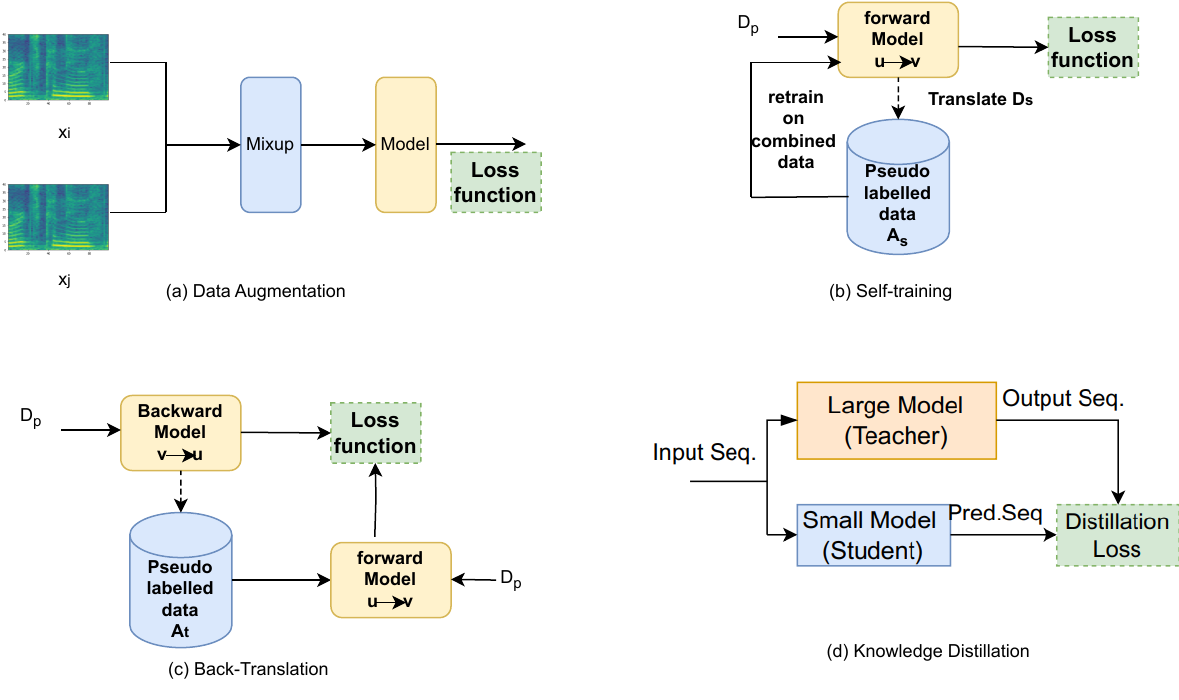}
    \captionsetup{font=small}
    \caption{Strategies for addressing data paucity in ST task modelling. (a) Data augmentation, (b) Self-training, (c) Back-translation, and (d) Knowledge distillation. The dashed arrow indicates that the model is used for inference.}
    \label{fig:dataaug}
\end{figure*}

\subsubsection{Augmenting speech data}
Speech data can be augmented in various ways. For example, by adding noise, speed and pitch perturbation, time and frequency masking to name a few. SpeechAugment \citep{Park2019SpecAugmentAS}   policy consists of warping the features, masking blocks of frequency channels, and time steps. It has been successfully used both for ASR \citep{vincent2017analysis} and ST tasks \citep{Bahar2019OnUS}. MixSpeech \citep{Meng2021MixSpeechDA} as shown in Fig. \ref{fig:dataaug}(a) takes the weighted combination of two different speech features as input and two recognition losses with the same weights.  A generalization of MixSpeech \citep{Xie2023MixRepHR} called MixRep applies the mixup idea to the acoustic feature and hidden layers inputs.  MixRep combination with a regularization term along the time axis further improves ASR performance. Both MixSpeech and MixRep have been shown to perform well for low-resource ASR and their effectiveness is still to be tested for ST tasks. M3ST \citep{Cheng2022M3STMA}  applies two levels of Fine-Tuning (FT) using mixup data-- word, sentence, and frame level mix data in the first FT level and source speech and transcription mixup in the second FT level. M3ST achieves SOTA on MuST-C compared to baselines.

\subsubsection{Augmenting speech and text data}
It is possible to augment both speech and text simultaneously and create new paired data. For example, sample, translate, and recombine \citep{Lam2022SampleTR} first samples a  \emph{suffix replacement} from \emph{suffix memory} corresponding to a \emph{pivot} token from transcription. It then translates the combined new utterance (prefix+pivot+replacmenet suffix) to generate a new target sentence. The corresponding audio pair is obtained by concatenating the audio frames of the prefix, pivot, and replacement suffix.  The interesting thing about the proposed method is that it can generate \emph{real-looking} sentences contrary to \emph{pseudo-sentences}. \textbf{Concatenation} of original ST data has been used to augment the entire training data \citep{Lam2022MakeMO}. In particular, \citep{Lam2022MakeMO} proposes \emph{CatSpeaker} that uses single speaker information and \emph{CatRandom} that randomly generates audio-text pairs spoken by different speakers.

\subsection{Pre-training}
Pre-training is an approach to handle data scarcity for low-resource problems and is deemed as a form of transfer learning \citep{bozinovski1976influence}.  Data used for pre-training may consist of either speech, text, or both. Once the models are pre-trained leveraging augmented data, it enhances the robustness of the model on downstream tasks. We find that SOTA ST models often use pre-training on a large amount of ASR/MT corpus. In ST, pre-training has been used by many researchers \citep{paulik2013training, bansal2017towards, anastasopoulos2018tied, wang2020curriculum, Dong2021ListenUA,zhang2022revisiting, tang2022unified}. Pre-training has been applied in two flavors by different researchers: Independently and  Jointly.

In {\bf independent} pre-training, individual modules (encoder, decoder, semantic decoder, etc.) are pre-trained using auxiliary data such as ASR and MT data. Such an approach has been followed by \citep{wang2020curriculum, chen2020mam, zheng2021fused}. In particular, \citep{wang2020curriculum} pre-trains the encoder using ASR data for learning semantic concepts.  \citep{chen2020mam}  propose a self-supervised method called Masked Acoustic Modeling (MAM), which randomly masks part of the speech spectrogram and then recovers it on top of the encoder. Whereas \citep{zheng2021fused} unifies speech and text representation through masked language modeling. 
Besides pre-training the encoder and the decoder, various researchers also exploit pre-trained feature extractors such as Wav2vec \citep{schneider2019wav2vec} used by \citep{Zhang2023TuningLL} and \citep{liu2020multilingual} HuBERT \citep{hsu2021hubert} used by \citep{Zhang2023DUBDU}. Very recently, \citep{Tsiamas2024PushingTL} proposed an ST model that pre-trains the speech encoder using optimal transport and CTC. They claim to surpass supervised ST models requiring no paired speech-text data in a \textbf{zero-shot} setting.  

In {\bf joint} pre-training, the entire model is first pre-trained in an E2E fashion followed by fine-tuning over the ST corpus \citep{Fang2023BackTF, Bapna2021SLAMAU}. It is often accompanied by multitasking pre-training with ASR, MT, and masked language modeling tasks \citep{Chung2021w2vBERTCC}, using supervised as well as unsupervised speech and text data. The  \citep{tang2022unified}  pre-trains on speech/text-to-text/speech, text-to-text, speech self-supevised learning (SSL), and speech-to-phoneme. SpeechT5 \citep{ao2021speecht5} pre-trains on ASR, ST, text-to-speech, speech conversion, and speech enhancement tasks. Wave2Seq \citep{Wu2022Wav2SeqPS} pre-trains jointly using pseudo-languages. Multi-modal multi-task pre-training leverages five tasks:  self-supervised speech-to-pseudo-codes (S2C), phoneme-to-text (P2T), self-supervised masked speech prediction (MSP),  supervised phoneme prediction (PP),  and ST task \citep{Zhou2022MMSpeechMM}.

\subsection{Self-training and Back-translation}
Both the Self-Training and Back-translation (BT) methods are approaches employed to harness monolingual data for training models that necessitate supervised data but encounter limitations in the availability of a sufficient supervised parallel corpus, as illustrated in Fig.\ref{fig:dataaug}(b) and (c). The self-training method is utilized to make use of source monolingual data, while the back-translation method is applied to target monolingual data. In the end, both methods are employed synergistically to generate augmented data.

More specifically, given a speech-text parallel corpus  $D_p=\{({\bf u}^i, {\bf v}^i)|i=1,2,\ldots, n\}$, monolingual source speech corpus $D_s=\{{\bf u}^i_s|i=1,2,\ldots, m\}$ and monolingual target text corpus  $D_t=\{ {\bf v}^i_t|i=1,2,\ldots, p\}$, where $m, p >> n$. In self-training, first, a translation model $f_{u \rightarrow v}$ is trained on $D_p$. It is then used to generate \quotes{pseudo labels} ${\bf v}^i_s$ for $D_s$ by applying $f_{u \rightarrow v}$ leading to auxiliary data $A_s = \{ ({\bf u}^i_s, {\bf v}^i_s)|i=1,2,\ldots, m\}$ . The combined data $D_p \cup A_s $ is then used to re-train the model  $f_{u \rightarrow v}$. Whereas in back-translation, $D_t$ is translated using a backward translation model $f_{v \rightarrow u}$ creating auxiliary data 
$A_t = \{ ({\bf u}^i_t, {\bf v}^i_t)|i=1,2,\ldots, p\}$ for training the forward translation model $f_{u \rightarrow v}$ on the combined data $D_p \cup A_t$. 

Back-translation on discrete units to train a unit-to-text translation model is applied in \citep{Zhang2023DUBDU} which is on par with methods leveraging large-scale external corpus. \citep{Fang2023BackTF} proposes a back-translation strategy for target-to-unit and unit-to-speech synthesis for low-resource language translation without transcript. \citep{Wang2021LargeScaleSA} extract speech features using wav2vec 2.0 pretraining, a single iteration of self-training and decoding with a language model. Cyclic feedback from MT output is used as a self-training mechanism for a cascade of ASR-MT model that shows how to exploit the direct speech-translation data in \citep{Lam2020CascadedMW}.

\subsection{Knowledge distillation}
Knowledge Distillation(KD) transfers learned knowledge from a large ensemble model (called teacher) to a smaller single model (called student) as shown in Fig.\ref{fig:dataaug} (d) \citep{Hinton2015_KnowledgeDistilation}. This process encompasses both model compression \citep{Bucilǎ_2006_Model_Compression} and transfer-learning. More details of recent works utilizing KD approaches for ST tasks are given in \S \ref{endtoend} (ST with MT) and \S \ref{stviabridge}.

\section{Segmentation and Representation Learning}\label{segrepr}
E2E ST models rely on segmented inputs because handling long inputs is a challenging task \citep{kim2017joint, Tsiamas2022SHASAO}. 
Segmentation is the problem of splitting the long speech/text sequence into smaller and more manageable segments whose representations can be learned. This section will shed some light on the segmentation and representation issues and offer some advice on how to tackle them. 

\subsection{Segmentation Learning}
As discussed above, segmentation is an important issue while building ST models. Segmentation of text is easy--they can be split based on the strong punctuation. This is what current MT models rely on. Similarly, ASR models give lower importance to segmentation due to the small local context window required for the task. The cascaded ST model can perform segmentation by applying ASR followed by monolingual translation to restore the lost punctuation followed by segmentation on them \citep{Matusov2007ImprovingST, Matusov2018NeuralST}. On the other side, the E2E ST models require sophisticated segmentation of the speech primarily due to the importance of out-of-order word relation between the input and output that exist as well as the absence of linguistic features.

Traditionally, segmentation of speech is done manually. Due to the cumbersome task, segmentation learning is warranted. Segmentation is done based on either \emph{length} which splits the speech at fixed-lengths or \emph{pause} which splits the speech based on Voice Activity Detection (VAD) \citep{Sohn1999ASM}. The third approach to segment the speech is \emph{hybrid} mode in that length and linguistic contents are taken into account \citep{Potapczyk2020SRPOLsSF, Gaido2021BeyondVA, Tsiamas2022SHASAO}. The hybrid approach surpasses the length and pause-based approaches to segmentation in terms of performance \citep{Gaido2021BeyondVA}. Concretely, \citep{Tsiamas2022SHASAO} learns the manual segmentation using a binary classifier and  \emph{probabilistic divide-and-conquer} algorithm \citep{Gaido2021BeyondVA} is used at inference time to decide the split point. However, there is still a gap in the hybrid and manual approaches to segmentations, and future work may consider paying attention to this.

Our discussion above focuses on segmentation in the \emph{offline E2E models}. Segmentation of speech in \emph{streaming E2E models} is presented in \S \ref{stinstreaming}.
\subsection{Representation Learning}
Representation learning is a type of machine learning where algorithms are supposed to discover and extract useful features automatically from the raw data. It has been successfully applied in computer vision \citep{Wu2020DeepRL}, natural language processing \citep{DBLP:journals/corr/abs-2102-03732}, and speech \citep{Mohamed_2022}. Representation learning is an important issue in ST tasks because speech and text are two distinct modalities of data that reside in different embedding spaces. Hence, we not only need better representation learning methods for speech and text but also their joint representation learning. Many of the works in ST apply speech/text representation learning methods before actually applying encoder-decoder or transducer-based methods (explained later in \S \ref{endtoend}) for the ST task. Below, we provide details of such representation learning methods used for ST tasks.
\subsubsection{Text  Representation}
ST models often use ASR transcripts and MT translations as auxiliary data which needs to be fed to the encoder and decoder, respectively. To learn representation for such text data, existing works rely on word embedding \citep{zhang2023improving, berard2016listen}, LSTM \citep{kim2017joint, weiss2017sequence, berard2018end, jia2019leveraging}, and Transformer \citep{Wang2021LargeScaleSA, liu2021cross, Zeng2021RealTranSES}. Text data is often tokenized and fed as either a word or as a character \citep{berard2018end}. The output of the decoder could be graphene, characters, or words.

\subsubsection{Speech Representation}
 ST models take speech as input and utilize various speech-based feature representation methods to convert speech into a vector representation. Traditional speech feature extraction methods such as Perceptual Linear Prediction,  (PLP), Fbank, and Mel-Filter Cepstral Coefficient (MFCC) \citep{rabiner2010theory} have been used after normalization to extract speech features by many \citep{duong2016attentional,berard2016listen, kim2017joint, berard2018end, anastasopoulos2018tied, bansal2018pre, jia2019leveraging, Inaguma2019MultilingualES,liu2020bridging, Dong2021ListenUA, Le2023PretrainingFS, parcollet2024lebenchmark}, sometimes combining them with pitch features and speech augmentation methods as described in \S \ref{dataissues}.  These feature extraction methods are sometimes being replaced by distributed feature representation methods such as speech word2vec \citep{chung2018speech2vec} owing to their dense continuous feature representation capability. 
 
 It is difficult to get a large amount of labeled speech data to learn supervised speech feature representation. Therefore, more recent works exploit speech features learned via unsupervised and self-supervised ways, mapping continuous speech signal to \emph{discrete units}-- akin to words and sub-words in the text domain. Such a representation facilitates tools developed in NLP to borrow in the speech domain.  Among them the most popular is  Wav2Vec \citep{schneider2019wav2vec} and its variants such as w2v-BERT \citep{Chung2021w2vBERTCC} and Wav2vec 2.0 \citep{Baevski2020wav2vec2A} used in \citep{Tran2020CrossModalTL, Le2020,   Li2020MultilingualST, Han2021LearningSS, popuri2022enhanced, zhang2023improving}. Interestingly, Wav2Vec and its variants can be used as an encoder in a Seq2Seq framework alone or combined with adapters and CNN for \emph{Length shrinking}\footnote{Length shrinking is an important issue in ST task since speech is a much longer sequence than text. Therefore, existing works employ various techniques such as \emph{length adapters, CNN, CTC} for length shrinking}. A few works such as CSTNet \citep{khurana2020cstnet, wang2020curriculum} use CNN for feature extraction and length shrinking.

 More recent works in ST are employing HuBERT \citep{hsu2021hubert} for speech representation (among other benefits of HuBERT) \citep{Zhang2023DUBDU}. Hubert offers stable training and better targets than Wav2Vec 2.0 since it uses hidden layers representation during the clustering process. For encoding long-speech signals, Conformers \citep{Gulati2020ConformerCT} can be used as they provide local context through convolution block and global context through an attention mechanism. Seamless4MT \citep{Communication2023SeamlessM4TMM} uses conformer for speech encoding.
 
 Other speech representation techniques such as VQ-VAE \citep{Oord2017NeuralDR}, WavLM \citep{Chenwavlm2021}, data2vec \citep{baevski2022data2vec}, Robust data2vec \citep{zhu2023robust}, SpeechLM \citep{zhang2024speechlm},  may also be explored while encoding speech for ST tasks.

\begin{figure}
    \centering
    \includegraphics[height= 4cm, width=4cm]{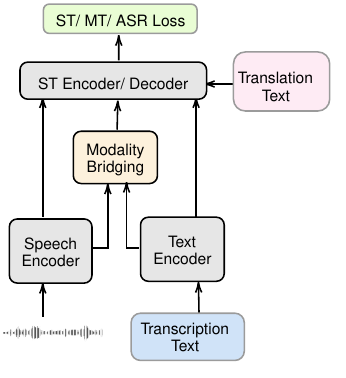}
    \captionsetup{font=footnotesize}
    \caption{Modality Bridging}
    \label{fig:modbridge}
 \end{figure}
\subsubsection{Joint Speech-Text Representation} \label{stviabridge}
The speech and text in an ST task are semantically related because both of them refer to the same thing. Therefore, it is imperative to learn a joint speech-text representation in the hope of bridging the modality gap between them. A method for learning a combined representation of text and speech is called {\bf modality bridging} (see fig.\ref{fig:modbridge}).  Hence, a good ST model should learn a representation such that embeddings of both modalities for similar speech-text pairs lie close to each other.  It is believed that low performance on ST tasks is due to models not learning aligned representations of speech and text.  Therefore, different authors have devised different ways to fill the gap, which fall into five major approaches: (a) adapters, (b) contrastive learning, (c) knowledge-distillation, (d) optimal transport, and (e) mix-up strategy. Below we discuss the works utilizing these approaches and show the pros and cons.
\begin{enumerate}[(a)]
    \item  {\bf Adapters} are small modules integrated with pre-trained networks for specific tasks \citep{houlsby2019parameter}. They have performed on par with fine-turning-based approaches while requiring only a fraction of trainable parameters. For example, in \citep{Gllego2021EndtoEndST, 
 zhao2022m, sarkar2023direct}, the modality gap is filled using \emph{adapter layers}, which is a \emph{multi-headed self-attention} with pooling operation. The author uses Wav2Vec 2.0 \citep{Baevski2020wav2vec2A} for speech-feature extraction, wherein self-attention layers in the transformer are equipped with pooling operation for dimensionality reduction to match the text representation. 
 \item {\bf Contrastive learning} approximates the \quotes{semantic} distance in the input space using a simple distance in the target space after mapping input patterns onto the target space \citep{chopra2005learning}. It tries to bring positive instances closer while pushing negative ones apart. It has been used excessively in both supervised and unsupervised settings for learning representations. For example, \citep{zhang2023improving} performs the \emph{explicit knowledge transfer} through contrastive learning. It learns \emph{frame} and \emph{sentence-level speech} feature representation and uses \emph{whitening} \citep{su2021whitening} to alleviate the  MT representation degeneration.  \citep{liu2019end} decouples the encoder representation into three parts: acoustic encoder, shrinking (done via CTC) of acoustic encoder output, and semantic encoder for modality-gap bridging.  Using a contrastive learning architecture, Chimera \citep{Han2021LearningSS} trains a \emph{semantic memory module} which is shared for overcoming the modality distance. XSTNet \citep{ye2021end} augmented with contrastive loss \citep{ye2022cross} investigates three different methods: \emph{span masked representation, word-repetition} and \emph{cut-off}.  It claims that contrastive loss is better than CTC and L2 loss. \emph{Word-aligned contrastive learning} (WACO) \citep{ouyang2022waco} bridges the modality gap by forming average speech and word embedding of the same word as the positive pair while of different words as negative pairs. CSTNet is a \emph{self-supervised} learning framework based on contrastive learning (using a mix of triplet losses)\citep{khurana2020cstnet}. 
On top of the CTC loss, the boundary-based speech length shrinking mechanism is applied in \citep{Zeng2022AdaTranSAW}. The authors claim that if \emph{boundary-based shrinking} is applied with other modality-bridging techniques, such as contrastive loss, it can further improve the model performance. The approach presented achieves lower inference speed and memory footprint.  \citep{yin2023improving} proposes a novel integration of speech and text, referred to as a third modality. This fusion is achieved through the application of Cross-modal Contrastive Learning \citep{sohn2016improved} and Cross-Attentive Regularization \citep{tang2021improving}. Additionally, the method incorporates techniques such as Knowledge Distillation and Jensen-Shannon Divergence \citep{lin1991divergence, liu2019end, gaido2020end} to bridge the modality gap, addressing challenges related to input representation, semantics, and hidden states.
\begin{table*}
  \centering
  \resizebox{\textwidth}{!}
  {
  
  \scriptsize
  \begin{tabular}{p{4cm}p{3cm} p{2cm} p{2.5cm} p{1.5cm} p{1.5cm} p{1.5cm} p{1cm}}
    \toprule
    Models/Techniques & Problem Solved & Dataset & Language Pair & Speech (hours) & Metric (BLEU)\\
    \midrule
       M-Adapter + W2V2 + mBart \citep{Baevski2020wav2vec2A} & training gap between Pre-training \& Fine-tuning the modality & MuST-C & En$\rightarrow$De & 408 & 25.9 \\
       &  &  & En$\rightarrow$Ro & 432 & 24.62 \\
       &  &  & En$\rightarrow$Fr & 492 & 37.34 \\
       
       \hline
       
       Chimera \citep{Han2021LearningSS} & projecting audio \& text to a common semantic representation & MuST-C & En$\rightarrow$De & 408 & 27.1 \\
       &  &  & En$\rightarrow$Fr & 492 & 35.6 \\
       &  &  & En$\rightarrow$Ru & 489 & 17.4 \\
       &  &  & En$\rightarrow$Es & 504 & 30.6 \\
       &  &  & En$\rightarrow$It & 465 & 25.0 \\
       &  &  & En$\rightarrow$Ro & 432 & 24.0 \\
       &  &  & En$\rightarrow$Pt & 385 & 30.2 \\
       &  &  & En$\rightarrow$Nl & 442 & 29.2 \\
       \hline
       ConST (XSTNet + Constrastive Loss) \citep{ye2021end} & closes modality gap & MuST-C & En$\rightarrow$De & 408 & 28.3\\
       &  &  & En$\rightarrow$Es & 504 & 32.0 \\
       &  &  & En$\rightarrow$Fr & 492 & 38.3 \\
       &  &  & En$\rightarrow$It & 465 & 27.2 \\
       &  &  & En$\rightarrow$Nl & 442 & 31.7 \\
       &  &  & En$\rightarrow$Pt & 385 & 33.1 \\
       &  &  & En$\rightarrow$Ro & 432 & 25.6 \\
       &  &  & En$\rightarrow$Ru & 489 & 18.9 \\
       \hline
       W2V2 + mBart + Adapter \citep{Gllego2021EndtoEndST, zhao2022m} & slow convergence speed & MuST-C & En$\rightarrow$De & 408 & 28.22 \\
       WACO \citep{ouyang2022waco} & limited parallel data (1-hour) & MuST-C & En$\rightarrow$De & 1 & 17.5 \\
       AdaTrans \citep{Zeng2022AdaTranSAW} & closing gap between length of speech \& text & MuST-C & En$\rightarrow$De & 408 & 28.7 \\
       &  &  & En$\rightarrow$Fr & 492 & 38.7 \\
       &  &  & En$\rightarrow$Ru & 489 & 19.0 \\
           
       \hline
       STEMM \citep{fang2022stemm} & Speech representation & MuST-C & En$\rightarrow$De & 408 & 28.7 \\
       &  &  & En$\rightarrow$Fr & 492 & 37.4 \\
       &  &  & En$\rightarrow$Ru & 489 & 17.8 \\
       &  &  & En$\rightarrow$Es & 504 & 31.0 \\
       &  &  & En$\rightarrow$It & 465 & 25.8 \\
       &  &  & En$\rightarrow$Ro & 432 & 24.5 \\
       &  &  & En$\rightarrow$Pt & 385 & 31.7 \\
       &  &  & En$\rightarrow$Nl & 442 & 30.5 \\        
       \hline
       CTC loss + Optimal Transport (Siamese-PT) \citep{Le2023PretrainingFS} & without change in architecture & MuST-C & En$\rightarrow$De & 408 & 27.9 \\
       &  &  & En$\rightarrow$Es & 504 & 31.8 \\
       &  &  & En$\rightarrow$Fr & 492 & 39.2 \\
       &  &  & En$\rightarrow$It & 465 & 27.7 \\
       &  &  & En$\rightarrow$Nl & 442 & 31.7 \\
       &  &  & En$\rightarrow$Pt & 385 & 34.2 \\
       &  &  & En$\rightarrow$Ro & 432 & 27.0 \\
       &  &  & En$\rightarrow$Ru & 489 & 18.5 \\
       \hline
       Fine \& Coarse Granularity Contrastive Learning \citep{zhang2023improving} & limited knowledge transfer ability & MuST-C & En$\rightarrow$De & 408 & 29.0 \\
       &  &  & En$\rightarrow$Fr & 492 & 38.3 \\
       &  &  & En$\rightarrow$Ru & 489 & 19.7 \\
       &  &  & En$\rightarrow$Es & 504 & 31.9 \\
       &  &  & En$\rightarrow$It & 465 & 27.3 \\
       &  &  & En$\rightarrow$Ro & 432 & 26.8 \\
       &  &  & En$\rightarrow$Pt & 385 & 32.7 \\
       &  &  & En$\rightarrow$Nl & 442 & 31.6 \\
    \bottomrule
  \end{tabular}
  }
  \caption{\label{seq2se2_modality} Performance of the ST models using modality bridging. The datasets, language pairs, duration of speech, and metric(BLEU) are shown. }
\end{table*}
\item 
{\bf Knowledge-distillation} \citep{Hinton2015_KnowledgeDistilation} is a mechanism to \emph{distill} information from a trained and large \quotes{teacher} model to a smaller and efficient \quotes{student} model. It has been used with $L_2$ loss in \citep{huzaifah2023analysis} to address the modality gap issue.

\item {\bf Optimal transport} (OT) \citep{peyre2019computational} is a mechanism for comparing two \emph{probability distributions} .
In the ST task, speech and text representations may be deemed as two probability distributions, and therefore, OT can be applied. More formally, suppose $\alpha$ and $\beta$ denote the discrete probability distributions corresponding to speech and text representations. The masses at each position $u_i$ and $v_i$ are $a_i$ and $b_j$ respectively such  that $\sum_{i=1}^m a_i=1$ and $\sum_{j=1}^n b_j$. Suppose further that the cost of transporting a unit of mass from $u_i$ to $v_j$ is $c(u_i,v_j)$, where $c$ is some cost function such as Euclidean distance. Let $Z_{ij} \geq 0$ be the quantity of mass to be transported from $u_i$ to $v_j$ then the goal of OT is to move all masses from $\alpha$ to $\beta$ such that the following objective function is minimized

\begin{equation} \label{Wasserteindistance}
    \min_Z \langle C, Z \rangle, \qquad Z{\bf 1}_n = a, Z^T{\bf 1}_m = b, Z \geq 0
\end{equation}
In the above eq., $C$ and $Z$ are the matrices whose elements are $c_{ij}=c(u_i,v_j)$ and $Z_{ij}$, respectively.  ${\bf 1}$ denotes the vector of ones. In ST task, $c(u_i,v_j) = \|u_i-v_j\|_p$ for some $p \geq 1$. The loss corresponding to \eqref{Wasserteindistance} is called \textbf{Wassertein loss} optimizing which is costly. Hence an entropy-regularized upper-bound approximation is often optimized 
\begin{equation}
    \min_Z \{\langle C, Z \rangle-\lambda H(Z)\}
\end{equation}
where $\lambda$ is a regularization parameter and $H(\cdot)$ is the von-Neuman entropy matrix.

Recent works make use of the OT as presented above. 
For example,  \citep{Le2023PretrainingFS} uses optimal transport and CTC together to close the modality gap during the pre-training phase. They show significant gains in BLEU score when the ST model is fine-tuned without any external data compared to multitask learning. Similarly, \citep{Tsiamas2024PushingTL, tsiamas-etal-2023-speech} uses OT+CTC to align the speech-encoder representation space with the MT embedding space whereas \citep{Zhou2023CMOTCM} aligns the two representations via OT followed by cross-modal mix-up at the token level.
\item {\bf Mix-up strategy:} Speech-Text Manifold Mixup (STEMM) \citep{fang2022stemm} strategy uses speech embedding. It mixes embeddings of speech and text into the encoder-decoder of a translation model for bridging the modality gap under the \emph{self-supervised learning} framework. 
PromptST \citep{yu2023promptst} presents a linguistic probing learning strategy, referred to as \emph{Speech-Senteval}, inspired by the approach introduced by \citep{conneau2018you}. This strategy is implemented on the higher layer of the encoder within pre-trained ST models, specifically targeting the challenges associated with learning linguistic properties that these models often struggle with at the higher layers.

Table \ref{seq2se2_modality} presents the performance scores of ST models based on modality-bridging techniques. We can observe that mixup strategy achieves the highest BLEU score on En-De pair.  Whereas boundary-based speech length shrinking mechanism matches the score when combined with other modality-bridging techniques.
\end{enumerate}
{\bf Discussion}: The study finds that adapters can shrink the speech length as well as the modality distance between the text and speech representations while requiring a small number of trainable parameters. The contrastive loss is found to be better than CTC and $L_2$ loss for modality-bridging.  The boundary-based speech length shrinking combined with contrastive loss may improve the ST task performance. Finally, it is possible to build ST models requiring zero parallel ST data \citep{Tsiamas2024PushingTL}.

\section{End-to-End ST Models} \label{endtoend}
End-to-end models for ST as discussed previously are gaining traction comparably to cascade models. This section presents an overview of E2E models. We categorize them under two major E2E themes: framework-based and data-based.  The first category is further divided on whether the framework used is offline or streaming. The second category is based on the nature of the data.  The sub-categorization presented in the data-based section depends upon which component boosts the ST task performance, as claimed in the papers. As such, the demarcation is not strict, and there may be overlaps in the subcategories. In addition, our emphasis in the present review of existing works is highlighting the core contribution and limitations as claimed by the authors. That means we look for answers to the question: what is the main technical contribution of authors to solve the ST problem? Thus, wherever possible, we have limited the mathematical description and believe such details can be found in the related papers. We attempt to provide a succinct and clear picture of what works and what does not while addressing the ST problem.

\begin{figure}
 
    \centering
    \includegraphics[height= 3cm, width=4cm]{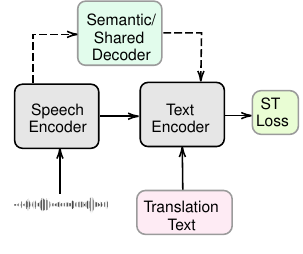}
    \captionsetup{font=footnotesize}
    \label{fig:figure1}
  \caption{E2E offline framework. The dashed arrow denotes optional components.}
  \label{fig:framework}
\end{figure}

\subsection{E2E ST Models based on Frameworks} \label{frameworks}
As mentioned in the previous section, E2E ST models based on the framework are further divided into whether the framework is offline or streaming. Below, we discuss both of these categories in detail.

\subsubsection{Offline Frameworks} \label{offline}
Offline frameworks perform ST tasks where output tokens are produced after having seen the entire speech utterance. These frameworks heavily rely on Seq2Seq architecture as shown in Fig. \ref{fig:framework}. It has an encoder for speech input, a decoder for text output, and an optional shared/semantic decoder connecting the encoder and the decoder. The model is usually optimized for the ST loss or sometimes in a \textbf{multitask learning} framework where ASR/MT/CTC \citep{GravesConnectionistTC} losses are combined with ST loss. Other times \textbf{Transfer learning} is utilized for leveraging pre-trained models for ST tasks. Another approach that has been gaining a lot of attention is \textbf{Non-Autoregressive modeling} (NAR) for the E2E ST task which gives faster inference. The following section will delve deeper into these approaches.

The Seq2Seq-based ST models proposed in the literature either use specialized encoders such as transformers or attention mechanisms which we discuss next.
\begin{enumerate}[(a)]
    \item {\bf Attention} mechanism is used to concentrate on specific sections of the input data instead of the entire data \citep{Larochelle2010LearningTC, Mnih2014RecurrentMO, Vaswani2017AttentionIA}. It has been a successful strategy for getting state-of-the-art (SOTA) results in NLP, computer vision, and other areas. 
There exist various types of attention in the literature such as \textbf{soft, hard, local, monotonic, multihead, self- and cross-attention, inter alia}. For more details, interested readers are encouraged to skim through \citep{Mnih2014RecurrentMO, Vaswani2017AttentionIA,  Brauwers2022AGS}. Below we provide efforts made to handle ST tasks using the attention mechanism within the Seq2Seq framework. 

 The \emph{convolutional} attention to \quotes{remember} and avoid translating the signal twice is used within Seq2Seq by \citep{berard2016listen}, which outperforms a hierarchical encoder with better results on {\it synthetic data} without using transcripts. The same author in \citep{berard2018end} uses source transcript and achieves results close to cascade models on LibriSpeech data. In \citep{duong2016attentional}, the author proposes phone-to-text alignment with a \emph{structural bias} feature in the attention model. The measurement of alignment has been explored in \citep{anastasopoulos2016unsupervised}, which uses IBM's translation model as well as \emph{dynamic time warping}\footnote{dynamic time warping (DTW) is an algorithm for measuring similarity between two temporal sequences, which may vary in speed.}. Seq2seq with attention trained using \emph{multitask learning} achieves promising results in \citep{weiss2017sequence}. These models, however, struggle with \emph{noisy inputs} and \emph{long acoustic signals} \citep{kim2017joint}. They use a joint CTC-attention model  \citep{GravesConnectionistTC} trained through multitask learning by incorporating \emph{regularizers}. The author uses two decoders where the second decoder seeks {\it higher level representation (HLR)} from the first decoder besides the encoder via the attention mechanism. \emph{Attention-Passing Model} (APM)  \citep{sperber2019attention}, which only passes high-attention vectors from the audio encoder to the translation text for decoding demands a smaller amount of data for training.
\item {\bf Transformer} is the architecture based on multi-headed self-attention \citep{Vaswani2017AttentionIA} which produces contextualized representation of the input. Because of parallelization and contextual representation, transformers have outperformed RNNs on several NLP tasks. This entails us applying transformers for the ST task as well. Transformer-based Seq2Seq with \emph{attention} is proposed in \citep{cattoni2021must}. The architecture has a \emph{quadratic memory complexity}, which involves: (a) CNN to downsample the input, and (b) 2-D attention to address short-range dependencies of spectrograms. In \citep{alastruey2022locality}, the weight of some attention is avoided for speech tasks, hence decreasing the size of the attention matrix. The transformer encodes the speech features, thereby introducing \emph{local self-attention} with a suitable window size for each layer to reduce the computational complexity. Other transformer variants which reduce its quadratic complexity such as \emph{perceivers} \citep{DBLP:journals/corr/abs-2103-03206} have been used as an encoder \citep{Tsiamas2022EfficientST}. Besides quadratic complexity, transformers require \emph{lossy} downsampling of speech features thus potentially throwing useful linguistic information. To tackle such issues, Speechformers have been proposed \citep{Papi2021SpeechformerRI} which aggregates information at higher layers based on more informed linguistic criteria.
\end{enumerate}

As discussed earlier, \textbf{multitask learning} combines the optimization of ST loss with an auxiliary loss such as ASR/MT/CTC loss. Another direction that has been explored by ST researchers is \textbf{transfer learning} in that Seq2Seq encoder/decoders are first \emph{pre-trained} using ASR/MT data respectively and then the entire model is \emph{fine-tuned} using ST data. Below, we discuss works based on multitask/transfer learning frameworks.

\begin{enumerate}[(a)]
    \item {\bf ST with ASR:} ST with ASR models make use of the transcript data along with speech-text pairs for pre-training. For example, curriculum pre-training \citep{wang2020curriculum} refers to using ASR data for pre-training a Seq2Seq model, allowing it to learn transcription. The author argues that if the model is further pre-trained on learning \emph{semantic concepts} (via  \emph{frame-based masked language modeling}) and \emph{word alignment} (via \emph{frame-based bilingual lexical translation)}, it boosts the ST task performance. Specifically, existing E2E models either pre-train the encoder or use multi-task learning for ST tasks. As such, the encoder cannot isolate the learning of three tasks: transcription, semantic concept, and alignment, which are segregated by \emph{dividing the labor}, and experiments prove the theoretical claims. Listen, Understand, and Translate (LUT)  \citep{Dong2021ListenUA} uses the Seq2Seq model with \emph{external} ASR loss. Their primary contribution is to introduce a \emph{semantic encoder network}, whose task is to use the encoder's output from transcription to minimize the mean-squared loss between the semantic representations and the BERT embeddings of the target text. Such a strategy implicitly builds and trains an NMT model for translation. Pre-training using ASR and/or MT has also been found useful in low-resource scenarios \citep{zhang2022revisiting}.


  

\item {\bf ST using MT:} \label{stusingmt}
This section discusses approaches that use either MT data for pre-training or directly using a pre-trained MT model in the ST decoder. These approaches rely on the idea of generating {\it pseudotext} and then translating them using MT. For example, Unsupervised Term Discovery (UTD) \citep{bansal2017towards}  groups repeated words into pseudo-text, which is subsequently used for training an MT model using the parallel pseudo-text and target translations. The main advantage of such a system is that it can translate some content words under low-resource settings. The overall results are not very promising on the Spanish-English Call-Home dataset. Another limitation of this work is that the approach is \emph{not} an E2E in a true sense as it involves two models-- a UTD and an MT model. A weakly supervised learning method for ST \citep{jia2019leveraging} that outperforms multi-task learning takes advantage of the pre-trained MT and TTS synthesis module. Pre-trained MT model is used as a \emph{teacher} to guide the \emph{student} ST model in \citep{liu2019end} (such an approach is dubbed as \emph{knowledge distillation} (KD)).  They, however, rely on \emph{source language} text and do not improve upon the pipeline system. Following along, \citep{Gaido2020OnKD} explores \emph{word, sentence} and \emph{sequence-interpolation} based KD approaches for transferring knowledge from pre-trained MT to ST model. 

  

\item {\bf ST using both MT and ASR:} 
This section discusses works employing MT and ASR pre-trained models \citep{Bahar2020StartBeforeEndAE, tsiamas-etal-2022-pretrained} or losses for transfer or multitask learning.  

Multitask learning proves to be effective when CTC loss is combined with ASR and MT loss in \citep{bahar2019comparative} using various E2E ST architectures such as direct, multitask many-to-one, one-to-many, tied-cascade, and tied-triangle. They show that pre-trained models with ASR and MT losses achieve promising results. Contrary to claims of \citep{anastasopoulos2018tied}, \emph{tied-triangle} architecture is no better than a direct model when fine-tuned properly. Since the ST task is similar to the MT task from the output perspective, works such as XSTNet \citep{ye2021end} utilize \emph{external} MT data to pre-train the encoder-decoder network extensively, then fine-tune it using parallel corpus data of MT, ST, ASR, and external MT data for optimizing the model using what they call \emph{progressive training}.
They achieve impressive performance on MuST-C and augmented Librispeech data. They also demonstrate improved performance on auxiliary tasks of MT and ASR. STPT model \citep{tang2022unified} proposes four sub-tasks for multitask pre-training: text-to-text (T2T), which is self-supervised; speech-to-phoneme which is supervised; acoustic learning, which is self-supervised, and ST which is supervised. Only T2T and ST tasks would subsequently be used for fine-tuning. Despite pre-training on \quotes{unlabeled} speech data, they obtained superior results on MuST-C data for the ST task. COSTT \citep{Dong2020ConsecutiveDF} pre-trains encoder using ASR data, the decoder using paired MT data, and then fine-tunes for the joint transcription-translation task. ComSL is a composite ST model relying on multitask learning with three losses ($L_{ASR}, L_{MT}, L_{ST}$) combined with cross-modality loss to bridge the gap \citep{le2023comsl}. It is worth mentioning that ComSL does not require forced-align ST data and learns the cross-modality alignment during training. This however requires optimizing four different losses, viz. \emph{Masked Token Prediction, Speech to Text Mapping, Encoder Representation Matching, and Decoder Distribution Matching}\footnote{please  see \citep{le2023comsl} paper for more details.} similar to \citep{Tang2021ImprovingST}. Fused acoustic and text encoding-ST (FAT-ST) \citep{Zheng2021FusedAA} follows the similar pre-training and fine-tuning idea as ComSL except that they propose to use any combination of training data from $D_{2^{\{u,x,v\}}}$ \footnote{$2^{\{u,x,v\}}$ is the power set of triplets.}. Essentially, they rely on masked language modeling (MLM) and translation language modeling (TLM) for pre-training \citep{conneau2019XLM}.

\item {\bf Non-Autoregressive Modeling}\footnote{ We present the discussion of NAR within the multitask learning framework because all NAR E2E ST models are optimized within the multitask framework. }
As discussed in the background section, an alternative approach to Autoregressive (AR) modeling is Non-Autoregressive (NAR) modeling. AR assumes that the output tokens are conditional dependent on the previously generated tokens. However, it causes significant latency during inference. NAR models solve this problem 
 by outputting all the translated tokens in parallel thus speeding up the inference. Formally, they are given by \eqref{nonautoreg}
 \begin{equation}\label{nonautoreg}
    p({\bf v}| {\bf u}; \theta) =\prod_{t=1}^{T_v} p(v_t|, {\bf u};\theta) 
\end{equation}
  There has been a surge in applying the non-autoregressive models (AR) in ASR and MT and it has prompted ST researchers to apply to it too. For example, \citep{Inaguma2020ORTHROSNE, Inaguma2021NonautoregressiveES} trains NAR and autoregressive decoding conditioned on a shared speech encoder. Another line of NAR works \citep{Chuang2021InvestigatingTR} explores CTC with ASR as an auxiliary task. CTC-based encoder only architecture (\citep{Inaguma2020ORTHROSNE, Inaguma2021NonautoregressiveES} use encoder and decoder both) for NAR E2E ST task is shown to perform comparably or better than strong AR models in \citep{xu-etal-2023-ctc}.

\end{enumerate}

{\bf Discussion}: Our study of Seq2Seq-based frameworks for ST task reveals that (a) structural bias can be obtained by stacked/pyramidal RNN  and alignment smoothing, (b) regularizers such as transitivity and invertibility improves Character Error Rate, (c) HLR helps in transcription as well as translation, and (d) changing the self-attention of the encoder with a logarithmic distance penalty enhances translation, (e) Progressive training needs a huge data and training time to achieve superior results, and (f) multitask pre-training can be used to leverage unlabeled speech data. \citep{zhang2022revisiting} shows that ST models trained from scratch using only ST tasks perform on par with or surpass pre-trained models. To achieve such results, proposed best practices include a smaller vocabulary, a wider feedforward layer, a deep speech encoder with the post-layer norm, CTC-based regularization, and parameter-distance penalty. Pre-training is still useful in low-resource data regimes. Transferring knowledge via KD from pre-trained MT to ST causes gender bias, omission of sentences, and generic verbal-tense choice.   Use of {\bf large vocabulary and models} is effective for NAR E2E ST task \citep{Inaguma2020ORTHROSNE}. It indicates that leveraging NAR with LLMs may be a future direction to explore.

\subsubsection{Streaming frameworks}
\label{stinstreaming}
Streaming frameworks for ST tasks start outputting target tokens on seeing only partial inputs, that is, the translation of the input as soon as it arrives without waiting for the entire input. They are also known as Simultaneous ST (SimulST or SST)\footnote{Note that in MT literature, some works such as \citep{IranzoSanchez2022FromST} differentiate between \emph{Streaming} and \emph{Simultaneous} setting where sentences are treated independently from each other. However, in ST, we find that existing works make no differentiation between them. } \citep{GoldmanEisler1972SegmentationOI, Fgen2007SimultaneousTO, tsiartas2013high, grissom-ii-etal-2014-dont}. It finds application in online speech translation and video dubbing, to name a few. Traditionally, the streaming ST problem has been solved by feeding the segmented output of a streaming ASR model to a streaming MT model \citep{Oda2014OptimizingSS, iranzo2020europarl}. However, due to the cascade nature of the model, it is prone to high latency and error propagation \citep{Arivazhagan2019MonotonicIL, Arivazhagan2020RetranslationVS, Zaidi2022CrossModalDR}.  The SST problem faces several issues in practical implementation; {\bf reordering,  acoustic ambiguity, and variable speech rate, and long inputs} being prominent among them. Our literature survey reveals that most of the existing works focus on handling \emph{long streaming inputs} and therefore, the discussion underneath revolves around that. Other issues mentioned above may also be considered for designing practical SST models.  
 \begin{figure}
    \centering    \includegraphics[width=8cm, height=4cm]{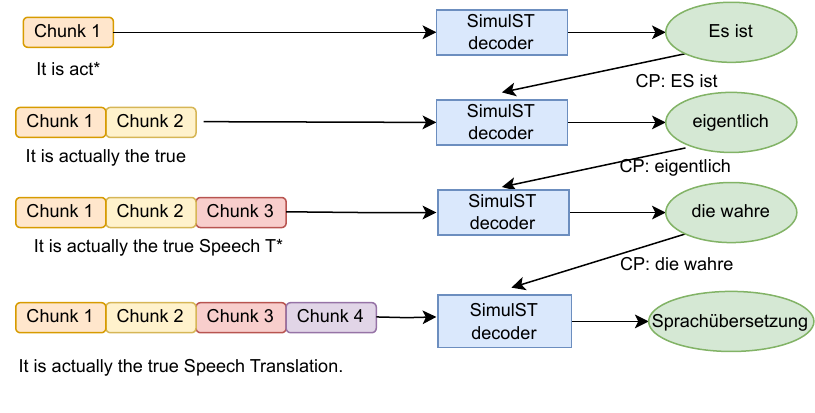}
    \caption{Incremental Decoding Framework. CP stands for the common prefix. Fig. adapted from \citep{Guo2024RBIRB}}
    \label{fig:Incremental Decoding Framework}
\end{figure}
 Existing streaming frameworks intervene Seq2Seq framework at various places to design SST models. These are (a) encoder-level, (b) decoder-level, and (c) input/latent-level. 
 \begin{enumerate}[(a)]
     \item {\bf Encoder-level:} SOTA SST models use transformers as encoders. Due to the \emph{self-attention} operation which looks at the entire utterance, it is unsuitable for streaming inputs. There exist some works that design encoders specialized for streaming inputs.  For example, \emph{augmented memory transformer} \citep{wu20i_interspeech, Ma2020StreamingSS}  splits the utterance $U$ into smaller-segments $S=[s_1,\ldots]$. Each segment $s_n$ consists of left context $I_n$, main context $c_n$, and right context $r_n$. Self-attention is calculated at the segment level only thereby reducing the time complexity. Augmented memory propagates the information from one segment to the other.
    \emph{Incremental transformer}  \citep{Zhang2020FutureGuidedIT} leverages a unidirectional encoder based on unidirectional-attention with future context masked for handling streaming inputs.
     \item {\bf Decoder-level:} Instead of modifying encoders, some works such as \citep{dalvi-etal-2018-incremental, Liu2020LowLatencySS,  nguyen21c_interspeech, Guo2024RBIRB} propose \textbf{incremental decoding} (see fig. \ref{fig:Incremental Decoding Framework}). In this framework, input speech is divided into fixed-size chunks and decoded every time a new chunk arrives. To avoid distractions from constantly changing hypotheses, selected chunk-level predictions are committed to and no longer modified. The decoding of
the next chunk is conditioned by the predictions committed. Instead of conditioning on all chunk-level predictions, a \emph{prefix} function is chosen to select a partial hypothesis because early chunks contain limited information \citep{Liu2020LowLatencySS}.  There exist several strategies for choosing the \emph{prefix} function. For example, Hold-$n$ and LA-$n$ \citep{Liu2020LowLatencySS}, SP-$n$ \citep{nguyen21c_interspeech} and  Regularized Batched Inputs (R-BI) \citep{Guo2024RBIRB}. Of these, Hold-$n$  either withholds or deletes the last $n$ tokens in each chunk,  LA--$n$ involves displaying the agreeing prefixes of $n$ consecutive chunks. SP-$n$ stands for \emph{shared prefix} of all best-ranked hypotheses. Contrary to these, RB-I applies various augmentations to input chunks to achieve regularization and SOTA results on the IWSLT SimulST task.
     \item {\bf Input/latent-level:}  Since speech input is too fine-grained, deciding when to READ and WRITE is challenging. The existing works introduce \textbf{pre-decision} module which segments the input speech at fixed-chunks \emph{(fixed)} or word-boundary \emph{(flexible)}. Similarly, READ/WRITE policy can be \emph{fixed} or \emph{adaptive} \citep{ma-etal-2020-simulmt}. Most research in SST concentrates on either improving speech encoding or pre-decision while relying on fixed policies such as wait-$k$. In this section, we discuss fixed and adaptive pre-decisions/policies.  These techniques are combined with Seq2Seq frameworks to devise streaming ST models.

\begin{figure}
    \centering
    \subfloat[]{ \label{fig:waitkk}
    \includegraphics[height= 2.8cm, width= 3.4cm]{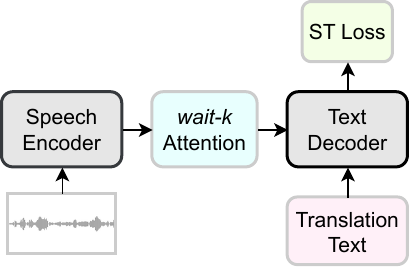}
    }
    \hspace{5em}
    \subfloat[]{ \label{fig:streaming}
    \includegraphics[height= 3cm, width= 3.3cm]{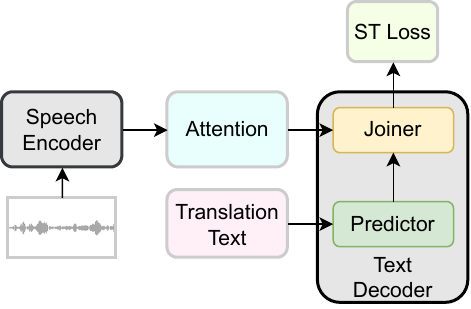}
    }
   
    \caption{ (a) wait-$k$ based Streaming ST (d) RNN-T based Streaming ST}
\end{figure}

{\bf Wait-$k$} policy \citep{Ma2018STACLST} (shown in fig. \ref{fig:waitk}) learns the parameters $\theta$ of the model by optimizing the negative log-likelihood $-\sum_{(\mathbf{u,v})\in D} \log p(\mathbf{v}|\mathbf{u}; k;\theta)$, where $k$ is the number of segments to look before starting translation (see Fig. \ref{fig:waitk}). The probability $p(\cdot)$ is calculated as

\begin{equation}
     p({\bf v}| {\bf u}; k; \theta) =\prod_{t=1}^{T_v} p(v_t|v_{<t},u_{t+k} ;\theta) 
\end{equation}
wait-$k$ policy guarantees that the model can look at $t+k-1$ speech segments while predicting token $v_t$ \citep{Ren2020SimulSpeechES}. However, one limitation of the wait-$k$ policy is that it fails to do a beam search while decoding except for long-tail \citep{Ma2018STACLST}. To solve this problem, \citep{Zeng2021RealTranSES} proposes a wait-$k$ stride-$N$ policy. It essentially is a wait-$k$ policy with the addition of $N$ READ and WRITE operations until the end of the sentence after reading the first $k$-segments. To determine the $k$-segments, \citep{Chen2021DirectSS} leverages streaming ASR to guide the direct simultaneous ST decoding via beam search. 
\begin{figure}
    \centering
     \includegraphics[height=4.5cm, width=6cm]{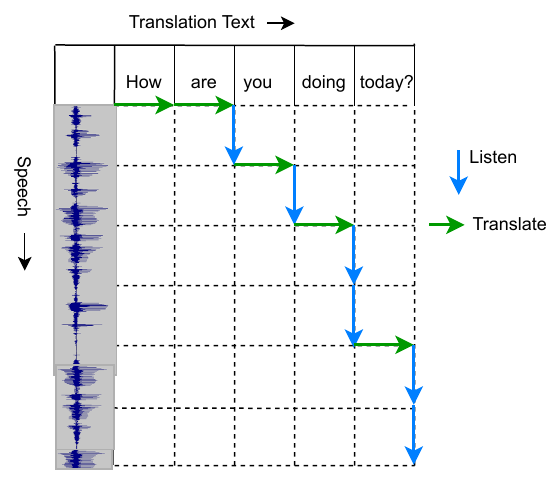}
    \caption{wait-$k$ strategy for streaming ST setting. In this, the decoder waits for $k$ input speech segments before starting to output. Thereafter, it produces one token for every source segment. The figure showcases the scenario with $k=2$. }
   \label{fig:waitk}
\end{figure}

As discussed above, determining when to write is crucial for efficient SST. Contrary to wait-$k$ policy which is a \textbf{fixed-policy}, {\bf Segmentation} can be performed on the embedded speech using CTC \citep{Ren2020SimulSpeechES}, attention mechanism \citep{Papi2022AttentionAA}, or incremental BEAM search \citep{Yan2023CMUsI2}. Essentially, these works \textbf{adapt offline ST to SST} showing spectacular performance on benchmark datasets. Note that the models proposed in \citep{Papi2022AttentionAA, Yan2023CMUsI2}  train models in a cascade manner while the inference is E2E.
Another issue with the fixed policy is that the model can not speed up or slow down appropriately with the input types. Other examples of \textbf{fixed-policy} are Wait-If* \citep{Cho2016CanNM} and Monotonic Chunkwise Attention (MoChA) \citep{Chiu2017MonotonicCA} that has been used in simultaneous MT and may be explored for SST.

The works mentioned above require that encoded speech be segmented so that the decoder can apply the wait-$k$ policy. The goal of segmentation is to identify the word, sub-word, or phone boundary which are usually not even (due to silences, longer syllables, etc.). That means the number of acoustic units varies with time in each segment. 
\emph{Monotonic-segmented Streaming ST} (MoSST) \citep{dong2021learning} is based on learning when to translate, which has a \emph{monotonic segmentation module} located between the acoustic encoder and the transformer. It has an {\bf Integrate-and-Fire} (IF) neuron \citep{Abbott1999LapicquesIO}, which fires above a threshold when the context is developed. If the context is not developed, the neuron receives signals and accumulates the acoustic vectors, thus mimicking \textbf{adaptive policy} for READ-WRITE operation. IF strategy has shown impressive performance in simultaneous ASR \citep{Dong2019CIFCI} and ST \citep{Chang2022ExploringCI}. It can be used for monotonic segmentation of the speech input along with adaptive decision strategy \citep{dong2021learning}. Another adaptive policy-based technique is Monotonic Infinite Lookback Attention (MILk) \citep{Arivazhagan2019MonotonicIL} used in simultaneous MT can be explored for SST. It essentially is a Monotonic Attention mechanism \citep{Raffel2017OnlineAL} that extends to infinite encoder states, theoretically, in the past and trains the MT model along with the MILk. It achieves better quality-latency trade-offs than MoCHA thanks to its soft attention to all the encoder states and hard attention. Monotonic Multihead Attention (MMA) \citep{Ma2019MonotonicMA} that extends MILK to multiple heads has been used for SST by \citep{ma-etal-2020-simulmt}. Its variants Efficient MMA \citep{Ma2023EfficientMM} solve numerical stability and biased monotonic alignment issues present in MMA but have not been explored for SST tasks. \textbf{Adaptive segmentation} based on an adaptive policy that takes into account acoustic features and translation history (called \emph{meaningful units}) is another effective mechanism for SST \citep{Zhang2022LearningAS}. 

Both fixed and adaptive policy mechanisms employ segmentation modules that are outside the translation module. As such, it breaks the acoustic integrity and potentially may drop the translation performance. Therefore, efforts such as \citep{Zhang2023EndtoEndSS} propose \textbf{differentiable segmentation} (DiSeg) learned jointly with the translation model using \textbf{expectation training}. DiSeg essentially predicts a Bernoulli random variable $\sigma(FFN(u_i))$, via a feed-forward network (FFN), to decide when to segment.  After segmentation, they apply segmented attention which combines unidirectional and bidirectional attention into one while masking future speech frames. Expectation training first constrains the number of segments followed by learning segmentation from the translation model both at semantic and acoustic levels \citep{Zhang2023EndtoEndSS}. 
 \end{enumerate}

The discussion so far covered encoder, and decoder level changes, and fixed and adaptive policies used for segmentation,  to develop SST models within the Seq2Seq frameworks. Another way to design SST models is by \textbf{Transduction}. It is the process of mapping a sequence to another sequence \citep{Jurafsky2008SpeechAL}. A transducer is a special type of Seq2Seq model that solves a few inherent problems. For example, online processing of the long inputs and monotonic sequence alignment is the biggest problem with Seq2Seq models \citep{Graves2012SequenceTW}, solved by transducers. Below we discuss a special type of transducer called {\bf RNN-T} and its improvements.

\begin{figure}
    \centering
    \includegraphics[width=\textwidth]{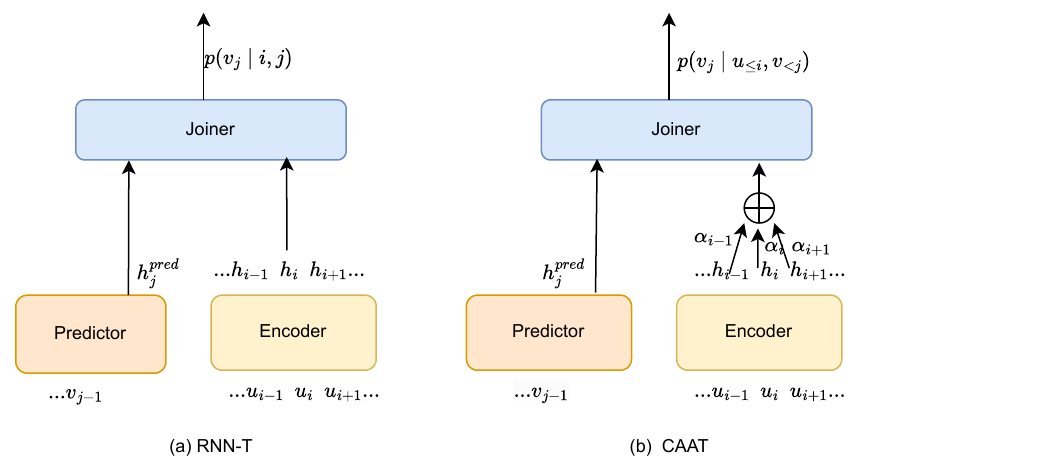}
    \caption{Architecture of (a) RNN-T (b) CAAT. Fig. adapted from \citep{liu2021cross}.}
    \label{fig:rnnt}
\end{figure}

 RNN-T is a transducer that can learn alignment between two sequences in an online/streaming fashion \citep{Graves2012SequenceTW} as shown in fig. \ref{fig:rnnt} (a). Formally, it learns the conditional probability $p({\bf v|u})$ by marginalizing all possible alignment paths $A(\bf u,v)$ including blank symbol $\phi$ as
 \begin{equation}
     p({\bf v|u}) = \sum_{{\bf \hat{v}} \in A(\bf u,v)}p({\bf \hat{v}|u}) 
 \end{equation}
 RNN-T differs from Seq2Seq in the sense that it divides the decoder into a \emph{predictor} and a \emph{joiner}. The predictor takes the previous time step output and yields the representation to be consumed by the joiner along with the hidden representation of the input from the encoder. Since the predictor does not look at the input, it can be pre-trained on the \emph{text-only} data in a low-data scenario. There have been several SST models proposed based on variants of RNN-T which we discuss next. 

 One of the main issues with RNN-T is the strict \emph{monotonic} alignment between the input and output sequences which makes them unsuitable for tasks requiring \emph{reordering} such as MT, ST, etc.  For example, \emph{Cross-Attention Augmented Transducer} (CAAT shown in fig. \ref{fig:rnnt}(b)) optimizes translation and policy models in tandem \citep{liu2021cross}. It eliminates the RNN-T's strict monotonic restriction for reordering in the translation. Using transformers as encoders to reduce the multi-step memory footprint causes a significant delay for CAAT. The use of regularization terms and substantial hyperparameter adjustment are some other limitations of CAAT. An extension of it in \citep{xue2022large} leverages \emph{Transformer Transducer} (TT) networks with attention pooling for streaming E2E ST tasks. \emph{Attention} divides the input audio into chunks of specific sizes. At any time, processing any input frame $\bf u_t$ can only see frames within its chunk and a fixed number of left chunks. By sharing the encoder, they also propose a variant to handle E2E ST tasks in {\bf multilingual} settings. The adaptive READ and WRITE policy choices between encoder output and ground truth contribute to its success. The same authors \citep{wang2022lamassu} propose to combine the benefits of \emph{language-specific} and \emph{language-agnostic} encoders within the TT framework. A shared encoder takes LIDs as gating values and computes weights for each language through the source LID scheduling scheme. The empirical results demonstrate superior performance and a smaller number of trainable parameters than bilingual ST. Adaptive (dynamic) policy for segmenting speech input has recently been explored in a Seq2Seq transduction setting by \citep{Tan2024StreamingST}. It essentially applies a cross-attention mechanism to decide when to segment the input followed by dynamic compression via \emph{anchor} representation. Thus, it saves memory and achieves a better latency-quality trade-off. 

Besides Transducer and Seq2Seq models, \textbf{re-translation} is another approach adapted for SST task by \citep{Niehues2016DynamicTF, Niehues2018LowLatencyNS, Arivazhagan2019ReTranslationSF, Arivazhagan2020RetranslationVS} though in a cascade setting. In this approach, the translated output can be re-generated after a fixed amount of time and displayed later for better quality.  Though it reduces latency by being greedy to display the partial translation, the output is highly unstable and causes \emph{flickring} effect. This may give rise to a bad user experience.  To mitigate instability, \citep{Arivazhagan2020RetranslationVS} propose a metric called \emph{erasure} which takes into the length of the suffix deleted during re-translation. Dynamic masking of MT output in a cascade of streaming ASR and MT for improving stability has been explored in \citep{Yao2020DynamicMF}. Another approach to reducing instability is luminance contrast and the Discrete Fourier Transform used in \citep{Liu2023ModelingAI}.

{\bf Evaluation of SST models:} SST models in the literature have been evaluated using the quality and latency metrics presented in \S \ref{evaluation}. Often showing a trade-off between quality and latency. Most of the existing works attempt to balance the quality and latency ignoring the visualization and cognitive load on the viewer when displayed on a screen. Towards this end, \citep{Papi2021VisualizationTM} emphasizes considering visualization as a metric to be evaluated along with the latency and quality. However, little effort has been made in this direction by the SST community. Therefore, we wish to draw the researcher's attention to also consider visualization as an evaluation metric for SST models. Towards this end, \citep{Liu2023ModelingAI} propose tokenized alignment, word updates with semantic similarity, and smooth animation of live captions. They find that it leads to a reduction in fatigue, and distractions while increasing the viewer's reading comfort.

  
        

\textbf{Discussion}: SST is a challenging problem and in that, E2E SST poses a further impediment. Our findings suggest that using \emph{adaptive} policy significantly improves the latency-quality trade-off.   Learned policy mechanisms have been an ongoing research and adapting them for true long-form SST may open new possibilities. Exploring differentiable segmentation for long sequences is still tapped and requires more investigation.  Re-translation is found to be on par with or better than SOTA streaming models \citep{Arivazhagan2020RetranslationVS} under a very low revision rate. Such a finding alludes to considering re-translation in an E2E SST system design. 

\subsection{ST Models based on the Nature of Available Data} \label{stnature}
In the previous section, we provided an overview of the ST models based on the frameworks used. The present section provides readers with another perspective on E2E ST models. In particular, it discusses the E2E ST models categorized based on the nature of the data, such as data is low-resource, streaming, multilingual, etc. Given the specific challenges they pose, we believe such a categorization might be interesting to researchers. 

\subsubsection{ST in Low-Resource settings} \label{stinlr}
\begin{figure}
    \centering
    \subfloat[]{\label{fig:lrl}
    \includegraphics[height= 3cm, width= 3.2cm]{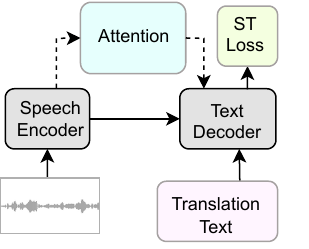}
    }
    \hspace{1em}
    \subfloat[]{ \label{fig:codemix}
    \includegraphics[height= 3cm, width= 3cm]{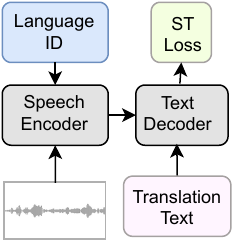}
    }
     \hspace{1em}
     \\
    \subfloat[]{ \label{fig:unsup}
    \includegraphics[height= 4cm, width= 4.2cm]{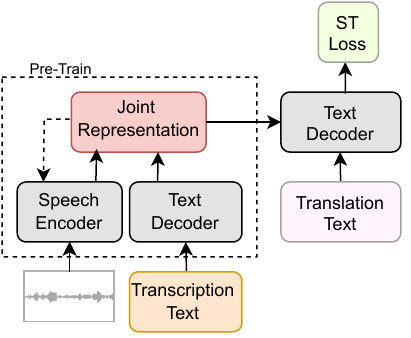}
    }
    \hspace{1em}
    \subfloat[]{ \label{fig:multilingual}
    \includegraphics[height= 4cm, width= 4cm]{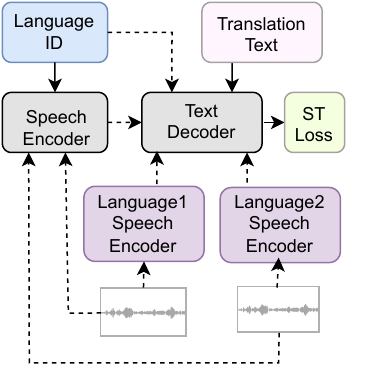}
    }
    \caption{E2E ST Models based on the nature of the data. (a) Low-resource ST, (b) Code-Mix ST,  (c) Unsupervised ST, and (f) Multilingual ST. The dashed arrow denotes optional components.}
    \label{fig:enter-label}
\end{figure}

A low-resource language (LRL) is one where speech and/or text data are scarcely available -- usually not enough to pre-train Seq2Seq models. As such, LRLs present challenges of their own such as overfitting and poor generalization. This section will discuss works where ST models are developed especially for low-resource languages. The proposed models under this category have generic architecture as shown in Fig.\ref{fig:lrl} which is similar to Seq2Seq ST models. We find the approaches mainly use pre-training the encoder on high-resource ASR data and subsequent fine-tuning on ST data. Another approach that has emerged in recent years to tackle LRL issues is \textbf{SSL}. For example, \citep{bansal2018pre} empirically demonstrates 100\% performance improvement on ST tasks. They find that if the ASR language differs from the source and target languages, then \emph{pre-training} on ASR data enhances ST task performance. Though the BLEU score is improved, the absolute BLEU score is only 7.1.  In \citep{wang2022simple}, the unsupervised ST is implemented for low-resource settings using pseudo-labels from unsupervised cascade models. SSL with \emph{discrete-speech unit} (DSU) has been used to fine-tune the ST model on limited ST data \citep{Lam2024CompactST}.

  



\subsubsection{Code-mix ST} \label{codemixedst}
Code-mix language refers to speech where one \emph{primary} language is used, but words or phrases from 
\emph{other (embedded)} languages are also included. This phenomenon arises from a multitude of challenges, encompassing ambiguous vocabulary, fluctuating lexical representations, intermingling of languages at the word level, redundancy, and alterations in word sequencing. Therefore, it is non-trivial to handle code-mixing while building ST models. 

We find that there exist only a few works on code-mix ST.
In \citep{weller2022end}, the code-mix dataset is created with the existing publicly available corpora Fisher \citep{cieri2004fisher} and Miami\footnote{\url{https://github.com/apple/ml-code-switched-speech-translation}}. As shown in Fig. \ref{fig:codemix}, code-mix ST models feed language ID in addition to speech input to the encoder of the Seq2Seq model \citep{weller2022end}. The Wav2Vec 2.0, an acoustic encoder, and mBART, a multilingual decoder, are used for both languages with an attention layer applied for the \emph{embedded} language.  The use of multilingual encoders and decoders is a common practice while building code-mix  ST models \citep{Yang2023InvestigatingZG}. In particular, self-supervised multilingual pre-training with adapters may be explored further. 



\subsubsection{Unsupervised ST} \label{unsupervisedst}
There is an abundance of unlabeled speech and text data. Since manual annotation and creating a parallel corpus is costly, the natural instinct is to exploit unlabeled data for training ST models. This section reviews works where researchers make use of the \emph{unlabeled} speech data to advance the ST task performance. 

For unsupervised ST tasks, it is common to leverage large-scale self-supervised and semi-supervised learning. For example, speech encoders such as Wav2vec 2.0 have been pre-trained in a self-supervised manner on Librilight data \citep{Kahn2019LibriLightAB} and used by  \citep{ Li2020MultilingualST, Wang2021LargeScaleSA} whereas the decoder is randomly initialized. The entire model is optimized on CoVoST 2 ST data, and the encoder is \emph{frozen}. Thereby, \emph{self-training} is executed to generate pseudo-labels for Libri-light data. The Wav2Vec 2.0 is a \emph{\quotes{student}} model which is fine-tuned with ground truth CoVoST 2 data and pseudo labels. Finally, a language model (LM) is trained on \emph{CommonCrawl} data and combined with the ST model to generate text via \emph{beam-search} decoding. Following along, for training the E2E model, \citep{Wang2021LargeScaleSA} produces pseudo-labels by cascading ASR, text \emph{de-normalization}, and MT in an \emph{Unsupervised} manner. Wav2Vec 2.0 and mBART are optimized for domain adaption using \emph{in-domain} data \citep{Li2020MultilingualST}. According to experimental results, the proposed method is effective for E2E models without pre-training. However, between supervised and unsupervised pre-trained models, performance gap is encountered, which may be investigated in future works. 



\subsubsection{Multilingual ST}\label{multist}
The multilingual ST model aims to translate from/to multiple speech input/output languages. It can be one of many-to-one, one-to-many, or many-to-many. The ST models solve multilinguality issues using mainly three approaches: (a) language ID, (b) dual-decoder, and (c) pre-trained models.

\begin{enumerate}[(a)]
    \item {\bf Language ID} (LID) is the \emph{identification label} that allows one to identify the target language and explicitly translate the speech simultaneously. The existing works handle \emph{multilinguality} using LID either with encoder or decoder. In \citep{Inaguma2019MultilingualES}, the model uses  LID in the decoder for one-to-many and many-to-many translation. They demonstrate impressive performance in translation from high-resource to low-resource languages without using any transcript data from LRL. However, using the LID embedding in the decoder \citep{Gangi2019OnetoManyME} is shown to underperform than using it in the encoder. The author shows that LID can be either \emph{concatenated} or \emph{merged} with the inputs and, when pre-trained with ASR data, can result in superior performance than the one-to-one system. The model, however, performs poorly when trained on many \emph{unrelated} target languages. One-to-many and many-to-one multilingual ST systems of \citep{Wang2020CoVoST2A, Wang2020CoVoSTAD} provide a good set of baselines for research purposes.
\item {\bf Dual-decoder} model is the transformer with two decoders, one for each ASR and ST, and the \emph{dual-attention} mechanism. In \citep{Le2020}, a dual-decoder model is proposed to optimize it for ASR and ST tasks jointly. The author hypothesizes that a dual-attention mechanism can benefit each task by transferring knowledge instantly or in \emph{wait-$k$} policy mechanism. Their model generalizes earlier models proposed for one-to-many and bilingual ST models. 
\item {\bf Pre-trained Multilingual Models} use a pre-trained encoder and decoder for acoustic modeling and language modeling, respectively. In \citep{Li2020MultilingualST, Tran2020CrossModalTL}, the author shows that efficiently fine-tuning \emph{mBART}, which is a pre-trained multilingual decoder \citep{Liu2020MultilingualDP} can achieve SOTA results on CoVoST data on \emph{zero-shot} cross-lingual and multilingual translation tasks.
Along similar lines, \citep{Le2021LightweightAT} shows that inserting \emph{adapters} in between layers of the encoder-decoder framework and tuning them can improve the ST task performance over bilingual ST models. SeamlessM4T \citep{Communication2023SeamlessM4TMM}, Whisper \citep{Radford2022RobustSR}, and other foundation models are built using many of these concepts like language ID in the decoder, multilingual, multimodal, and multitask pre-training.
\end{enumerate}
\subsection{Discussion}
The works presented so far show that E2E ST models have been improved tremendously. ST models' improved performance is likely due to leveraging pre-trained ASR/MT models or the respective corpus to train ST encoders/decoders. Weakly labelled/pseudo labels are another way to create more data for training ST models. Contrastive learning, mix-up strategy, adapters, and optimal transport are a few ways to bridge the modality gap. 

Applying unsupervised ASR and MT with the Wav2Vec 2.0 encoder and mBART decoder in a low-resource setting yields good results for ST models. When considering online data streaming, using the IF neuron for context building and translation improves results compared to using CAAT, which had latency issues due to reordering for translation tasks introduced by RNN-T. The mBART handles multilingual settings well by using a dual attention mechanism that facilitates knowledge transfer. Additionally, inserting adapters between the encoder and decoder layers improves performance. In the unsupervised ST setting, the SOTA results were achieved by training Wav2Vec 2.0 on data within the same domain as the speech. We see that the wait-$k$ policy is used in the streaming settings with segmentation and Multilingual settings with a dual-attention mechanism. In both cases, it yields good results. Also, {\it adapters} are used in modality bridging and multilingual settings with pre-trained models, which improves the performance. As shown in \citep{Sun2023TowardsAD}, multilingual E2E ST for LRLs can benefit when trained jointly with \emph{related} HRLs.


\subsection{Overall Performance Trend of E2E ST approaches in Common Benchmarks }
\begin{figure}
    \centering
    \begin{subfigure}[b]{0.4\textwidth}
        \centering
        \includegraphics[width=\textwidth]{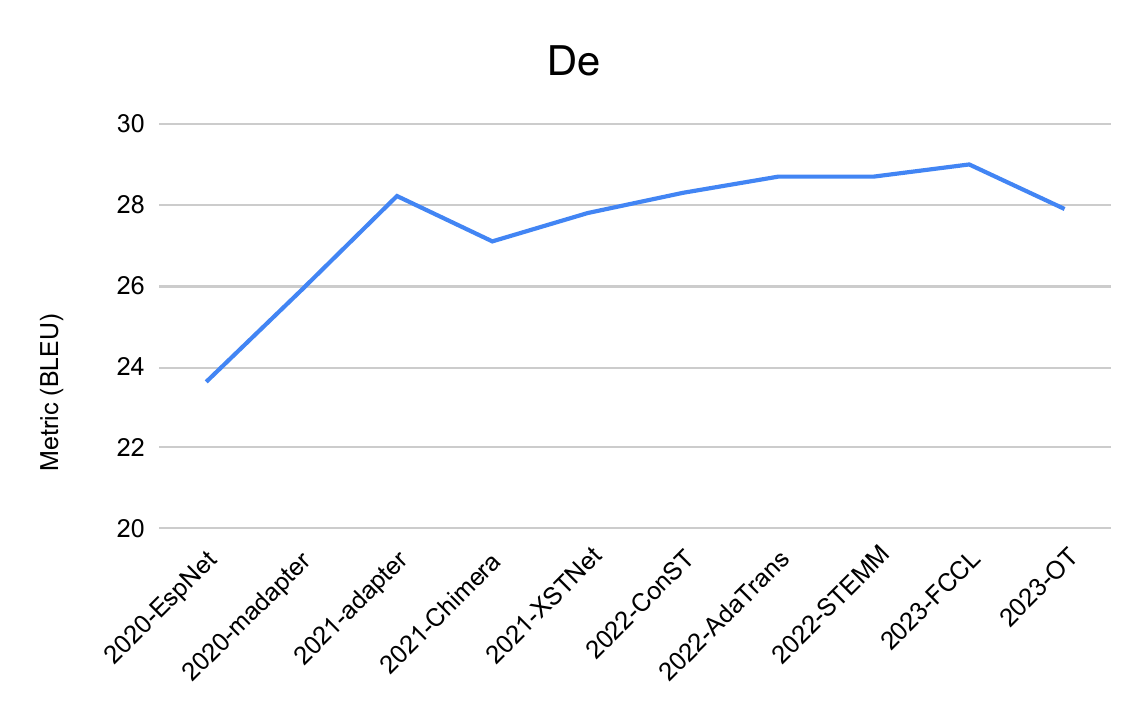}
        \caption{}
        \label{fig:enter-label}
    \end{subfigure}
 \begin{subfigure}[b]{0.4\textwidth}
        \centering
        \includegraphics[width=\textwidth]{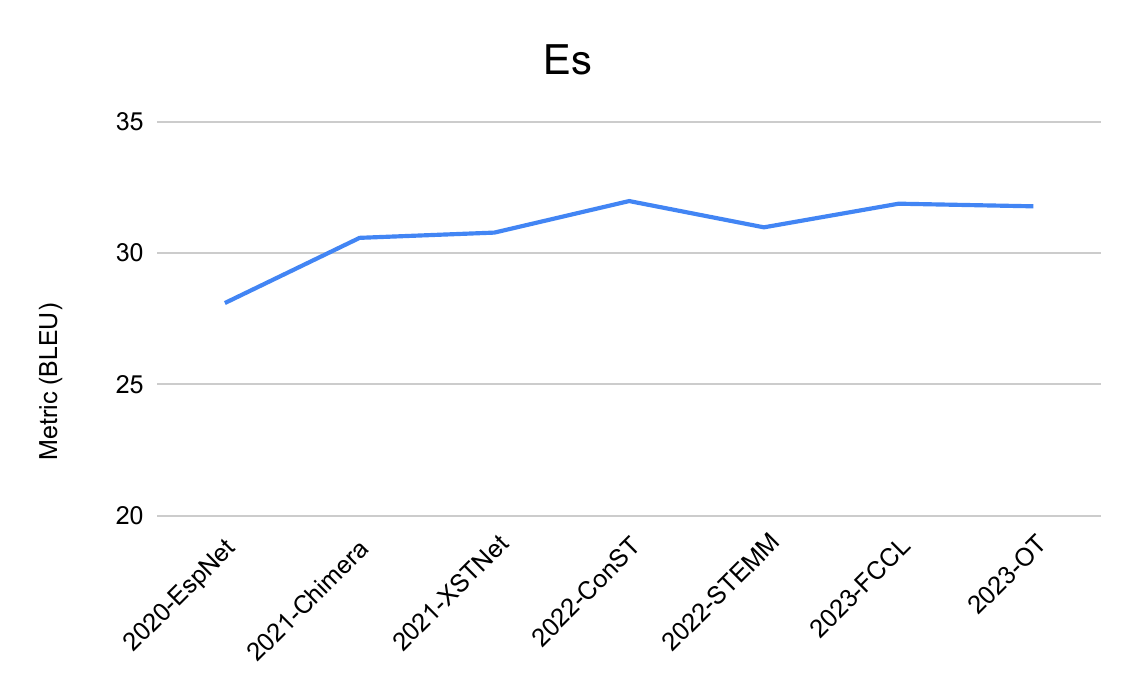}
        \caption{}
        \label{fig:enter-label}
    \end{subfigure}    
     \begin{subfigure}[b]{0.4\textwidth}
        \centering
        \includegraphics[width=\textwidth]{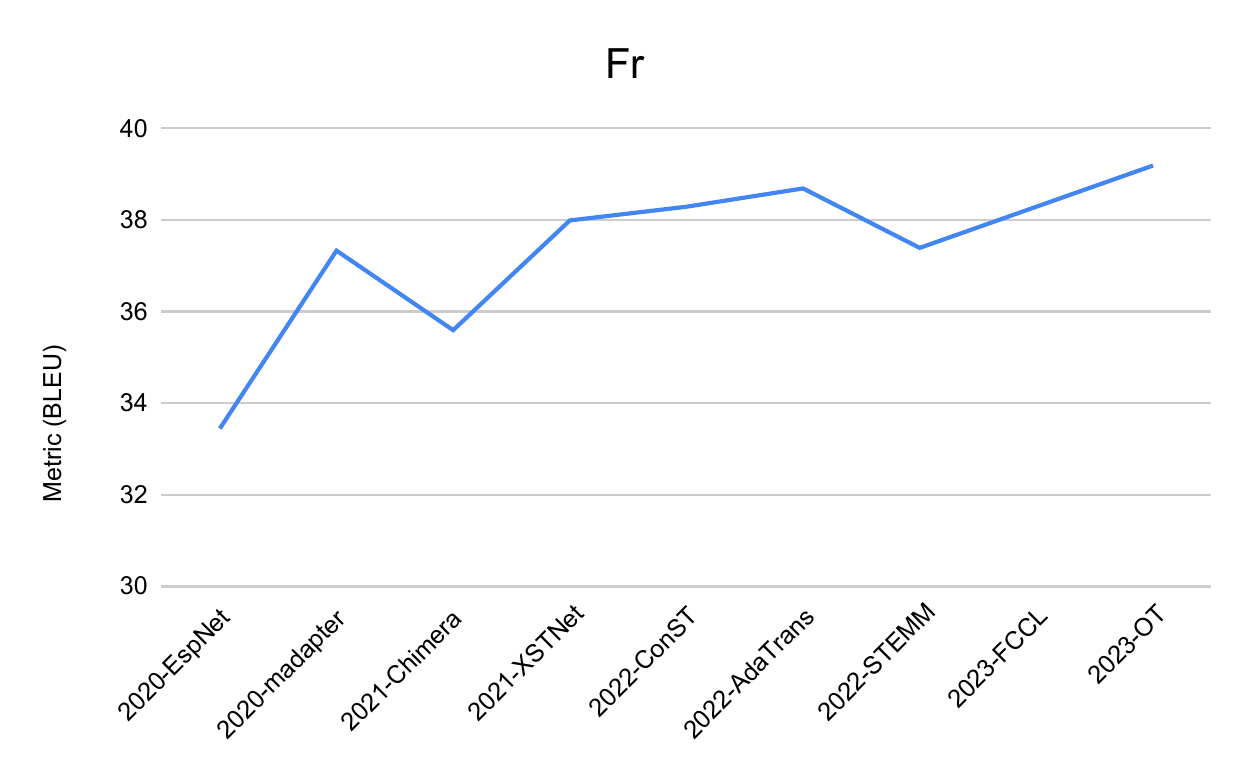}
        \caption{}
        \label{fig:enter-label}
    \end{subfigure}
 \begin{subfigure}[b]{0.4\textwidth}
        \centering
        \includegraphics[width=\textwidth]{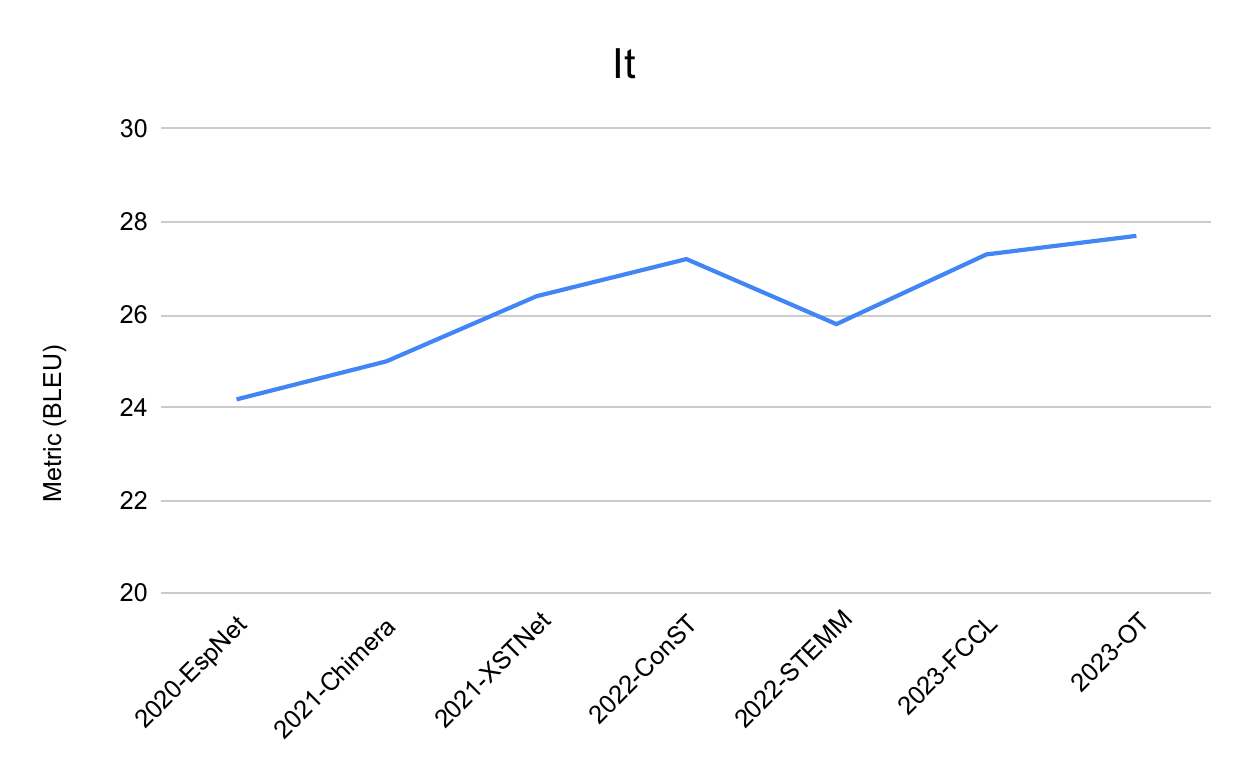}
        \caption{}
        \label{fig:enter-label}
    \end{subfigure}    
     \begin{subfigure}[b]{0.4\textwidth}
        \centering
        \includegraphics[width=\textwidth]{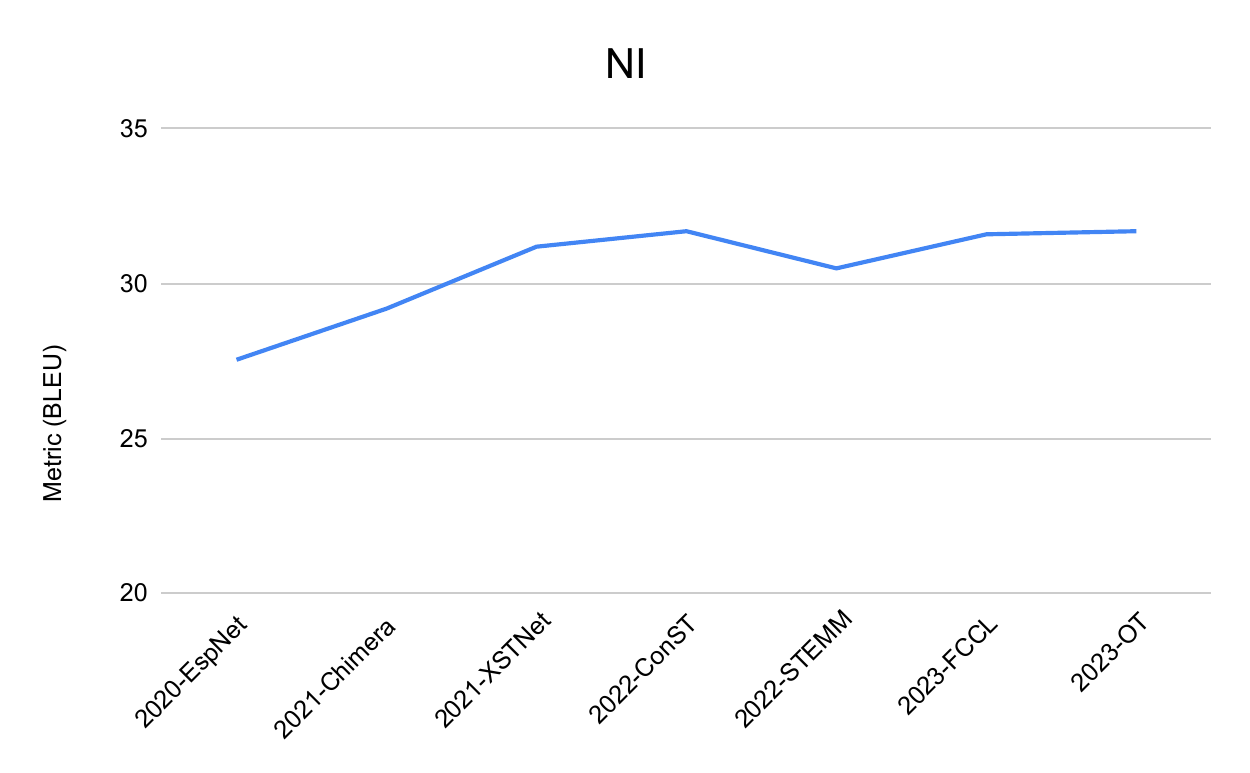}
        \caption{}
        \label{fig:enter-label}
    \end{subfigure}
 \begin{subfigure}[b]{0.4\textwidth}
        \centering
        \includegraphics[width=\textwidth]{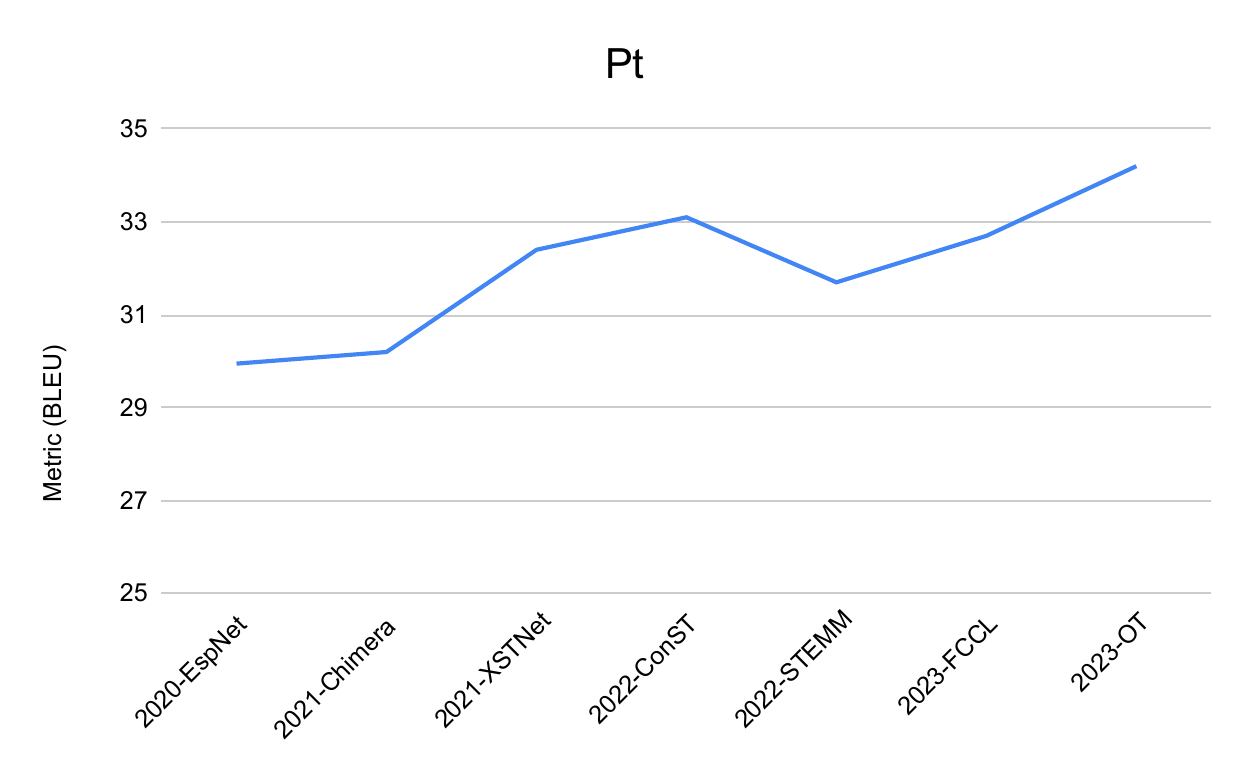}
        \caption{}
        \label{fig:enter-label}
    \end{subfigure}    
    \begin{subfigure}[b]{0.4\textwidth}
        \centering
        \includegraphics[width=\textwidth]{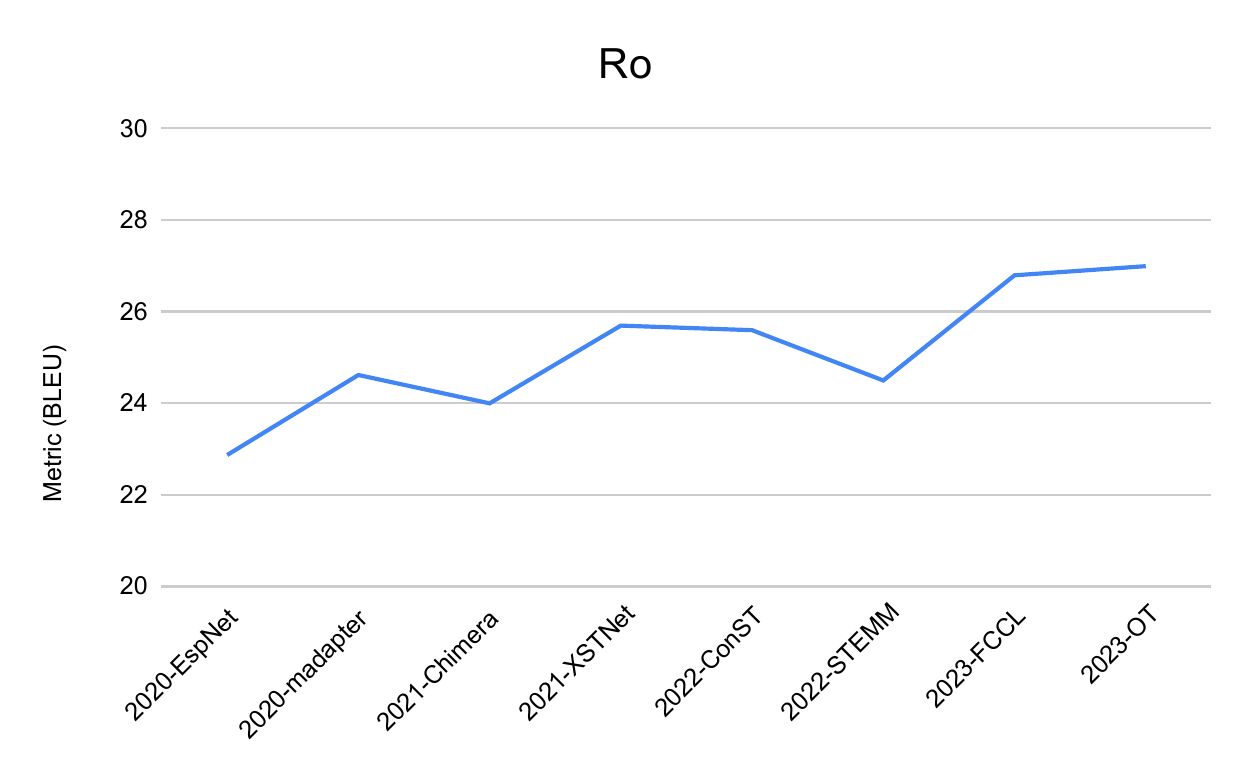}
        \caption{}
        \label{fig:enter-label}
    \end{subfigure}
 \begin{subfigure}[b]{0.4\textwidth}
        \centering
        \includegraphics[width=\textwidth]{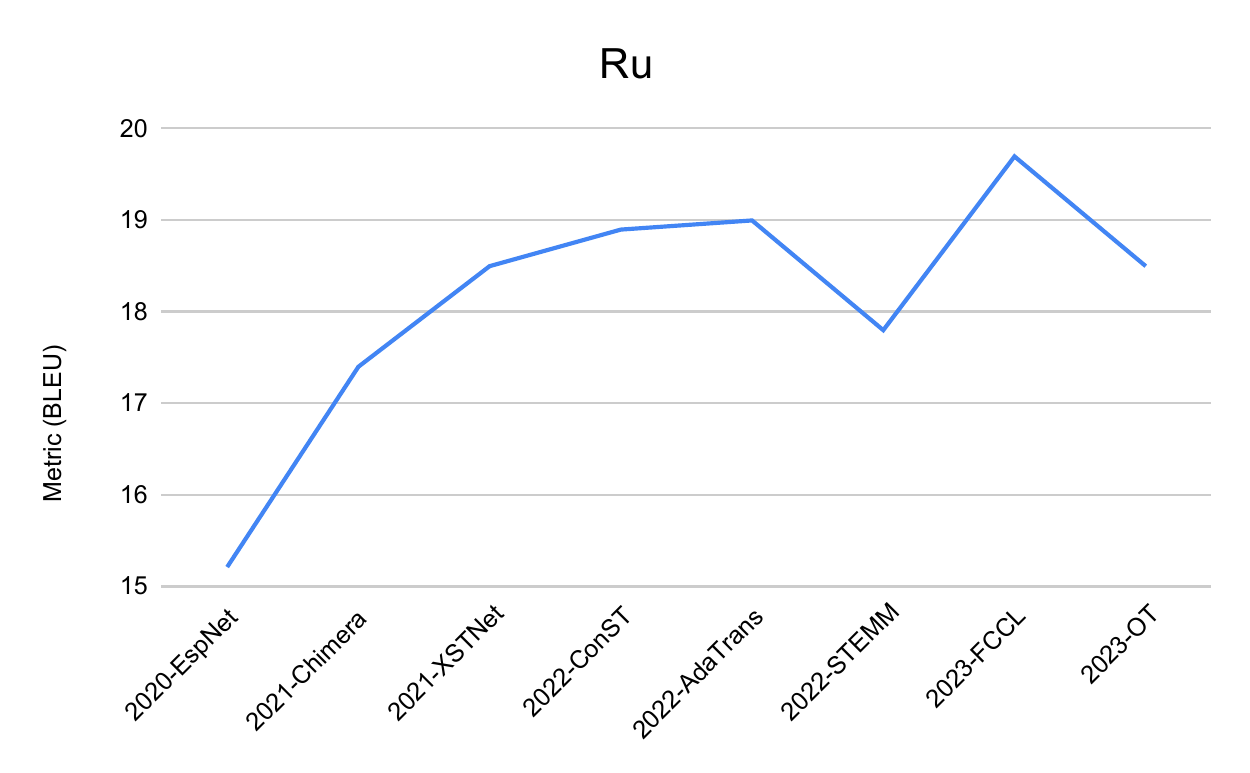}
        \caption{}
        \label{fig:enter-label}
    \end{subfigure}    
    \caption{Performance of ST models over MuST-C dataset}
        \label{fig:trend}
\end{figure}

In this section, we analyse the performance evolution of ST models over the MuST-C dataset, as depicted in Figure \ref{fig:trend}. We selected the MuST-C dataset due to its widespread adoption by researchers since its introduction in 2019. 

Figure \ref{fig:trend} reveals the overall performance of ST models over time has steadily improved across all 8 languages, with a few remarkable gains. The first significant gain was observed in 2021-adapter method \citep{Le2021LightweightAT}. This high jump in performance is achieved due to use of adapter layers within the \emph{multilingual} models that shows transferability of knowledge across related language pairs (note that not all proposed models tested their models across all 8 languages). It also shows that Chimera \citep{Han2021LearningSS}, which is a modality bridging model, performs poorly compared to adapter based models. That means, semantic shared network proposed in \citep{Han2021LearningSS} is not as good as adapters with multilingual models and there still is a gap between text and speech modality. 

The next jump we see is due to ConST \citep{ye2022cross} (for languages like Es, It, Pt, and Ru). This particular model achieved superior results by incorporating 
\emph{contrastive learning} to bridge the modality gap the first time. The cross-modal speech-text retrieval accuracy jumps from 4\% to 88\%! and better way to bridge the gap than Chimera. The drop in performance in STEMM compared to ConST is that both are from the same authors and were proposed in the same year. In fact, ConST is an improvement over XSTNet and STEMM by the use of cross-model contrastive loss. FCCL (medium model) \citep{zhang2023improving} further improves the performance, by applying contrastive learning over both the sentence- and frame-level, over ConST which applies contrastive learning only at the sentence level. Finally, OT based model outperforms contrastive learning based models on all languages except De and Ru. Looking closely, we find that OT based model \citep{Le2023PretrainingFS} is able to close the modality-gap only partially compared to ConST and FCCL for a few languages. Hence, as a recommendation,  coarse- and fine-grained contrastive learning and ASR pre-training with CTC loss via OT approaches may be explored to build better ST models. Note that LLM-based ST models are not compared here due to primarily their pre-training over massive amount of data and we want a fair comparison where pre-training over external ASR and MT corpus leads to higher performance as we find in ConST and FCCL models.


\subsection{SOTA Performance of E2E ST Models on Low-Resource Languages}
In Table \ref{LRL_Stats}, we present the SOTA performance of various ST models on low-resource language pairs as of November 2023. The table indicates which models, utilizing specific techniques, achieve SOTA performance. This provides a comprehensive overview of the current status of ST models for low-resource languages (LRLs). From Table \ref{LRL_Stats}, it is evident that the BLEU scores for many LRLs, such as Mn, Si, Ta, Id, Ja, and Sv, are relatively low. This is more likely due to small amount of speech data available for these  (as seen in Speech (hours) column)) compared to other LRLs where higher amount of speech data is used for training the LNA+Zero shot model. This highlights the need for improving the performance of ST models for these languages by increasing the data and designing better models.

\begin{table*}
  \centering
   \caption{\label{LRL_Stats} SOTA performance in Low-Resource Language Pairs: Dataset, Models, Speech Duration, Settings, and BLEU Score }
  \resizebox{\textwidth}{!}
  {  
  \scriptsize
  \begin{tabular}{p{2.5cm}p{4cm} p{2cm} p{2cm} p{3cm} p{1.5cm} }
    \toprule
    Language Pair & Model/Technique & Dataset & Speech (hours) & Setting & Metric (BLEU)\\
    \midrule
      
        Ainu$\rightarrow$En & Tied Multitask Learning with regularizers \citep{anastasopoulos2018tied} & Glossed Audio Corpus & 2.5 & ST with ASR \& MT &  20.3\\
        Mboshi$\rightarrow$Fr &  & Godard Corpus & 4.4 &  & 24.7 \\
        \hline
        Mt$\rightarrow$En & WACO \citep{ouyang2022waco} & IWSLT & 1 & Modality Bridging & 13.3 \\
        \hline
        Et$\rightarrow$En & Unsupervised + W2V2 + mBart \citep{wang2022simple} & CoVoST-2 & 3 & Low-Resource & 19.0 \\
        Lv$\rightarrow$En &  &  & 2 &  & 25.0 \\
        \hline
        En$\rightarrow$Ar & Teacher-Student (W2V2 + self-training + dec w/o LM) \citep{Kahn2019LibriLightAB}& CoVoST-2 & 430 & Unsupervised & 20.8 \\
        En$\rightarrow$Ca & &  &  &  & 35.6 \\
        En$\rightarrow$Tr & &  &  &  & 18.9 \\
        \hline
        Sl$\rightarrow$En & LNA + Zero Shot Learning \citep{Li2020MultilingualST} & CoVoST-2 & 2 & Multi-Lingual & 5.6 \\
        Sv$\rightarrow$En & &  & 2 &  & 5.9 \\
        Fa$\rightarrow$En & &  & 49 &  & 11.0 \\
        Tr$\rightarrow$En & &  & 4 &  & 11.2 \\
        Mn$\rightarrow$En & &  & 3 &  & 1.2 \\
        Ar$\rightarrow$En & &  & 2 &  & 6.4 \\
        Cy$\rightarrow$En & &  & 2 &  & 9.0 \\
        Ta$\rightarrow$En & &  & 2 &  & 0.9 \\
        Ja$\rightarrow$En & &  & 1 &  & 2.1 \\
        Id$\rightarrow$En & &  & 1 &  & 3.7 \\        
        En$\rightarrow$Cy & &  & 430 &  & 30.6  \\
        En$\rightarrow$Et & &  & 430 &  & 22.2\\
        En$\rightarrow$Fa & &  & 430 &  & 21.5\\
        En$\rightarrow$Id & &  & 430 &  & 29.9\\
        En$\rightarrow$Ja & &  & 430 &  & 39.3\\
        En$\rightarrow$Lv & &  & 430 &  & 21.5\\
        En$\rightarrow$Mn & &  & 430 &  & 14.8\\
        En$\rightarrow$Sl & &  & 430 &  & 25.1\\
        En$\rightarrow$Sv & &  & 430 &  & 30.4\\
        En$\rightarrow$Ta & &  & 430 &  & 17.8\\
        
    \bottomrule
  \end{tabular}
  }
 
\end{table*}

\section{Deployment of E2E ST Models}
Deployment of offline E2E ST models incurs several challenges. The first challenge is handling Cross-talk, noise, and background music removal and getting a clean speech. If the speaker is having stuttering, different dialect and accent then the same ST model may not work effectively. The second challenge is related to the distance of the speaker from the microphone and movements of the speaker around the microphone which can hamper the input speech quality. As a solution to these problems, the ST model may be trained over a variety of speakers in various acoustic conditions. The third challenge is related to memory consumption especially when considering LLM-based ST model deployment. To deploy memory-intensive and LLM-based ST models on edge devices, pruning, quantization, and knowledge distillation techniques may be used \citep{zhou2022deep} which significantly reduces the memory load.

Streaming ST models on the other hand are used as a submodule within the automatic subtitling. Hence their deployment has challenges of subtitling tasks which is considered harder. For example, subtitling requires the following challenges to be solved: (a) firstly, translated text should be segmented such that it reduces the cognitive load and maximizes the user experience like reading speech and synchronization with the speech (b) how many characters and lines to display? These constraints are usually decided by the media industries. For example, displaying a maximum of 2 lines of subtitles, 42 characters per line at max, and a maximum reading speech of 21 characters/second is used by TEDx \citep{agrawal-etal-2023-findings}. 

\begin{table*}
  \centering
   \caption{\label{dataset_Stats} Dataset statistics(\ding{51} means that feature is available for the dataset and \ding{55} means that the feature is unavailable for the dataset) }
  \resizebox{\textwidth}{!}
  {
  
  \normalsize
  \begin{tabular}{p{3cm}p{2.5cm} p{2.5cm} p{1.5cm} p{1.5cm} p{1.5cm} p{1cm} p{1cm}}
    \toprule
    \bf{Datasets}  & \bf{Source Language (Speech)} & \bf{Target Language (Text)} & \bf{Speech (hours)} & \bf{Speakers} & \bf{Validation} & \bf{Gender} & \bf{Age Group}\\ 
    \midrule
      MuST-C & En & 14 lang & 0.4K & 1.6K & \ding{55} & \ding{55} & \ding{55}\\
      
      Librispeech & En & Fr & 0.2K & 1.4K & \ding{51} & \ding{51} & \ding{51}\\
      
      CoVost & En & 11 lang & 0.7K & 11K & \ding{51} & \ding{51} & \ding{51}\\ 
      
      CoVost2 & 21 lang & En & 2.8K & 11K & \ding{51} & \ding{51} & \ding{51}\\ 
      & En & 15 lang & 0.7K & 78K & \ding{51} & \ding{51} & \ding{51}\\ 
      
      EuroparlST & 4 lang & 4 lang & 0.25K & \ding{55} & \ding{55} & \ding{55} & \ding{55}\\ 
      
      VoxPopuli & En & 15 lang & 1.79K & 4.3K & \ding{55} & \ding{55} & \ding{55}\\ 
      
      Kosp2e & Ko & En & 0.2K & 0.2K & \ding{55} & \ding{55} & \ding{55}\\ 
      
      GigaST & En & De, Zh & 10K & \ding{55} & \ding{55} & \ding{55} & \ding{55}\\ 
      Prabhupadavani & en-bn-sn code-mix & 25 lang & 0.09K & 0.13K & \ding{55} & \ding{55} & \ding{55}\\ 
      How2 &  En & Pt & 2K & \ding{55} & \ding{55} & \ding{55} & \ding{55}\\
      FLEURS & 102 lang & 102 lang & 1.4K & 0.3K & \ding{51} & \ding{51} & \ding{55} \\
     BSTC & Zn & En & 98 & \ding{55} &\ding{51} & \ding{55} & \ding{55} \\
     Indic-TEDST & En & 9 lang & 189 & 1.64K & \ding{55} & \ding{55} & \ding{55} \\
    \bottomrule
  \end{tabular}
  }
 
\end{table*}
\section{Resources for ST}\label{resources}
\subsection{Datasets for ST Tasks} \label{dataset}
There have been several datasets created for the ST task. Some of them are listed below, and we describe them here briefly.  Table \ref{dataset_Stats} provides information on various dataset statistics, such as hours of speech, the number of speakers, whether the dataset was manually or machine validated, the gender, and the age range to which the speaker belongs. Additionally, the tools required for creating these datasets are (a) {\it Gentle} \citep{ochshorn2017gentle} for audio-transcription alignment, and (b) {\it BertAlign}\footnote{\url{https://github.com/bfsujason/bertalign}} for transcription-translation alignment.
\begin{itemize}
    \item {\bf How2} \citep{sanabria2018how2} is an ST corpus of English instructional videos having Portuguese translations.
    \item {\bf Augmented Librispeech} \citep{kocabiyikoglu2018augmenting} is obtained from the LibriSpeech corpus \citep{panayotov2015librispeech}. It is a speech recognition repository generated using audiobooks of \emph{Gutenberg Project} \footnote{https://www.gutenberg.org/}. This dataset is designed to translate English speech into written French text. 
    
    \item {\bf CoVoST and CoVoST 2} \citep{Wang2020CoVoSTAD, Wang2020CoVoST2A}, the datasets are based on \emph{Common Voice} project \footnote{https://commonvoice.mozilla.org/en}. CoVoST is a many-to-one dataset covering 11 languages, while CoVoST 2 offers one-to-many and many-to-one translations for 15 languages.
    
    \item {\bf Europarl-ST} \citep{iranzo2020europarl} is a collection that contains speech and text data from \emph{European Parliament proceedings} between 2008 and 2012 in four languages. It includes multiple sources and targets for both speech and text.
     \item {\bf MuST-C} \citep{cattoni2021must} It is a large \emph{multilingual} ST translation corpus available . It contains translations from English into fourteen additional languages and is compiled from TED Talks. mTEDx \citep{salesky2021multilingual} is one such multilingual dataset from TED talks.  
    \item {\bf VoxPopuli} \citep{Wang2021VoxPopuliAL} dataset is an expansion of Europarl-ST. It includes data from European parliament sessions spanning from 2009 to 2020.
    
    \item {\bf Kosp2e} \citep {cho2021kosp2e} is a Korean (ko) to English(en) ST translation corpus, which contains Korean speech with parallel English texts. The corpus contains data from four different domains: \emph{Zeroth} from news/newspaper, \emph{KSS} \citep{park2018kss} from textbooks, emph{StyleKQC} \citep{cho2021stylekqc} from AI applications, and \emph{Covid-ED} \citep{lee2021building} from covid diaries of people which have emotions.
     \item {\bf BSTC} \citep{zhang-etal-2021-bstc} is a Baidu Speech Translation Corpus, a large-scale Chinese-English speech translation dataset. This
dataset is constructed based on a collection of
licensed videos of talks or lectures,  their manual transcripts, and translations into English, as
well as automated transcripts by an automatic
speech recognition (ASR) model. 
    \item {\bf GigaST} \citep{ye2022gigast} corpus is a collection of speech translations from English to German and Chinese. It is created using the English ASR GigaSpeech\citep{chen2021gigaspeech}, which features 10,000 hours of transcribed speech from various sources such as audioPortugesebooks, podcasts, and YouTube. 
    
    \item {\bf Prabhupadavani} \citep{sandhan2022prabhupadavani} is an ST dataset where speech is \emph{multilingual and Code-Mix} with three different languages, English is the primary language, and words and phrases from Sanskrit and Bengali are interjected. The text part has sentences in 25 languages.
    \item {\bf FLEURS} \citep{Conneau2022_FLEURS} FLEURS stands as a multilingual speech dataset, offering parallel recordings across 102 languages. Developed as an extension of the FLoRes-101 MT benchmark, it encompasses about 12 hours of annotated speech data for each language. 
   
\item {\bf Indic-TEDST} \cite{sethiya-etal-2024-indic-tedst} is a low-resource ST translation dataset across 9 Indic languages: Bengali (bn), Gujarati (gu), Hindi (hi), Kannada
(kn), Malayalam (ml), Marathi (mr), Punjabi (pa),
Tamil (ta), and Telugu (te).

\end{itemize}
Besides these popular ST datasets, there are some other \emph{smaller} size datasets such as {\it Fisher}\citep{cieri2004fisher}, {\it Call-Home}\footnote{\url{https://ca.talkbank.org/access/CallHome/eng.html}}, {\it Gordard Corpus}\citep{godard2017very}, {\it Glosse Audio Corpus}\footnote{\url{https://ainu.ninjal.ac.jp/folklore/en/}}, {\it BTEC} \footnote{\url{http://universal.elra.info/product_info.php?cPath=37_39&products_id=80}}, {\it WSJ}\footnote{\url{https://catalog.ldc.upenn.edu/LDC93s6a}}, {\it IWSLT}\footnote{\url{https://iwslt.org/}}, {\it Miami Corpus}\citep{deuchar2008miami}, and {\it MSLT Corpus} \citep{federmann2016microsoft}.

\subsection{Toolkits for ST} \label{toolkits}
To facilitate building and training ST models, various researchers have proposed a few toolkits. The toolkits for ST create an environment where the dataset for ST tasks can be pre-processed, and models can be trained, fine-tuned, and evaluated. We provide a short description of these toolkits to make the survey a place for a one-stop-shop for ST modeling.

\begin{itemize}
      \item {\bf SLT.KIT\footnote{\url{https://github.com/isl-mt/SLT.KIT}}}\citep{zenkel2018open} offers ASR, MT and ST models along with some specific features such as CTC and Attention based ASR, ASR with punctuation and a neural MT system.
    \item {\bf EspNet-ST\footnote{\url{https://github.com/espnet/espnet}}} toolkit \citep{inaguma2020espnet} is developed as there was no toolkit available for performing the sub-tasks of ST. EspNet-ST provides ASR, LM, E2E-ST, Cascade-ST, MT, and TTS along with examples. It also provided pre-trained transformer-based models on various datasets like MUST-C, Libri-trans, Fisher, CALL-HOME, and How2.
    
    \item {\bf FairSeq S2T\footnote{\url{https://github.com/facebookresearch/fairseq/tree/main/examples/speech_to_text}}} \citep{wang2020fairseq} toolkit is an extension to FairSeq\citep{ott2019fairseq} in which all the functions of EspNet-ST are available. Additionally, it provides the Non-Autoregressive MT, Online ST, and Speech Pretraining. The toolkit also provides state-of-the-art ST models based on RNN, transformers, and conformers. It has an in-built data loader for MuST-C, Librispeech, and CoVoST datasets. 
        
    \item {\bf NeurST\footnote{\url{https://github.com/bytedance/neurst}}} \citep{zhao2020neurst} is a lightweight toolkit, as it has no dependency on kaldi toolkit \citep{Zheng2011kaldi}. It has high computation efficiency using mixed precision and accelerated linear algebra and achieves faster training on large-scale datasets using Horovod \citep{DBLP:journals/corr/abs-1802-05799}.

\end{itemize}

  

\section{Future Directions for Research} \label{future}
This section highlights challenges that need the attention of researchers working on ST problems. 

\subsection{Cascade vs End-to-End Models}
As argued and presented through comprehensive experiments by \citep{Bentivogli2021CascadeVD}, the performance gaps between cascade and E2E ST models are bridged. However, as shown by \citep{agrawal-etal-2023-findings} in a recent IWSLT 2023 subtitling generation task, the performance of cascade models is far superior to E2E models for offline ST tasks evaluated on all metrics. Furthermore, as far as our understanding, no thorough assessment has been done for low-resource languages that use E2E and cascade models. It may be interesting to compare  E2E and cascade ST models on various ST datasets to assert the claims in the literature.
 
\subsection{ST on Code-Mix data}
We find that there exists limited study on the ST model that uses code-mix data as an input. A code-mix data has problems, such as different lexicons, syntax, and scarcity of labeled data. Therefore, it will be interesting to (a) create Code-Mix ST datasets incorporating more languages, (b) see how the existing ST models perform on code-mix ST data?, and (c) Can pre-training in many languages assist in tackling the code-mixing issue? 

\subsection{Domain-Invariant Models}
ST models developed for one domain do not scale well to other domains, as shown in the recent IWSLT 2023. Here domain in-variance setting is the ST model which is trained in some language combination (say Eng-De) and needs to be adapted to other language combinations (e.g., Eng-Hi). Transfer learning/continual learning can be explored to develop generic models. 

\subsection{Discrepancy between Automatic and Human Evaluation}
There may be discrepancies and disagreements among various metrics used to report ST task results. They do not match the mean option score (MOS) provided by human evaluators \citep{agrawal-etal-2023-findings}. For example, if a system evaluates the BLEU score between a ground truth sentence \emph{\quotes{Police shot the culprit with a gun}} and hypothesis sentence \emph{\quotes{Police use a gun to shot the culprit}}, it is 0! However,  both sentences above might be deemed appropriate translations of an utterance \emph{semantically} by an ST system. Such an argument is supported by dubbing artists who often change the voice of the sentence to simplify it or make it more pleasing.
\footnote{In the movie \quotes{Pirates of the Caribbean}, Jack Sparrow asks Bloom how long he can go for the girl. The original answer from Bloom is \quotes{I can die for her!}. Whereas Hindi dubbing is 
\quotes{Till the dying breadth} }

As highlighted in \citep{marie-etal-2021-scientific}, the BLEU score is being reported by more than 99\% of MT papers without accounting for statistical significance testing or human evaluation. Our survey of ST papers indicates the same trend being followed.  Therefore, we call for the attention of researchers to develop and use metrics that match human evaluations semantically.
An approach could be to subject the ground truth and hypothesis sentences under \emph{semantic textual similarity} tasks and score them accordingly.

\subsection{Handling Ambient Noise}
In our literature survey, we find that little has been done to deal with ambient noises. Ambient noise, background music, cross-talk, and non-verbal sounds may create difficulty in ST model learning. The model must distinguish between a meaningful utterance and ambient noise-- a non-trivial task. 

\subsection{Handling Multiple Speakers}
It is common in the real world where the audio/video has multiple speakers, each of which may have its own accent (cf., An Asian and American talking to each other in English), dialect, pitch, and accent. Performing \emph{speech separation} may be useful before feeding it to the ST model for improved performance. 

\subsection{Handling Speaker Diarization}
Speaker diarization refers to demarcating the timing of speakers in a multiple-speaker speech. So far, the datasets for ST do not have speaker boundary marks.
Creating speaker-diarized  ST data in a multilingual setting will be interesting to test the ST models' robustness.

\subsection{Multilingual and Simultaneous ST}
Multilingual ST has gained momentum recently due to its importance in the real world. For example, a single speech must be broadcast to multilingual communities (e.g., a conference is attended by a diverse group of people). It can be one-to-many, many-to-one, and many-to-many languages ST. Our literature survey shows that only a few works exist in this space. Besides, there is an opportunity to explore simultaneous multilingual ST, which is the most practical setting.

\subsection{Low-resource ST Datasets and Models}
Most existing works have focused on building ST models and datasets for high-resource languages. As we know, the success of ST models relies on the parallel speech-text corpus; building ST datasets for low-resource languages requires more attention. Further, a few works, such as \citep{bansal2018pre}, have reported ST task results on the Mboshi-French pair; however, the BLEU score is poor. Therefore, building models that transfer learning from language pairs with high to low resources is warranted. 

\subsection{LLMs  for ST tasks}
In the last few years, large language models (LLMs) have emerged as a promising solution to many NLP tasks including ST. LLMs show in-context learning (ICL) when trained over a massive amount of data. This process unlocks their hidden \emph{emergent abilities} \citep{Wei2022EmergentAO} and enables them for few-shot and zero-shot learning capability via prompting. There exist a few works \citep{Zhang2023TuningLL, Wu2023SpeechGenUT, Huang2023SpeechTW} (see \citep{Gaido2024SpeechTW} for a comparative discussion) which explore LLMs for ST task. Concretely, all of these models leverage a speech foundation model (SLM) followed by length adapter, modality adaptation, mixing the two modalities, and then LLMs for generating the output. GenTranslate \citep{Hu2024GenTranslateLL} builds upon the Seamless4MT by integrating an LLM on top
and performing $N$-best hypothesis tuning. Initial results are plausible. However, it remains to see how various components affect the downstream task performance, what is the best strategy for prompt design, and how to pre-train/fine-tune them in a parameter-efficient way for ST tasks. Further, the use of LLMs for SimulMT has been recently proposed \citep{Agostinelli2023SimulLLMAF} and it remains to see how to adapt SimulMT to SimulST.
\subsection{Really long Context Modelling}
As mentioned in the streaming section, SST models need to handle long input sequences. Current speech encoders lack infinite context modeling capability due to their quadratic complexity of self-attention. There have been recent improvements to handle the problem of infinite context. For example, Mamba \citep{zhang2024mamba}, Infini-attention \citep{munkhdalai2024leave}, and TransforerFAM \citep{hwang2024transformerfam} show some promising results in long context modeling. These models may be explored for SST task as well.

\section{Conclusion} \label{conclusion}
This survey paper delves into the most recent advancements in E2E ST translation works. Our discussion includes models, evaluation metrics, and datasets used to train ST models. We review various frameworks for ST models and highlight previous research in this field. The categorization of ST models is based on the kind of data they handle and the models employed. Additionally, we discuss potential future directions for improving speech-to-text translation. Our findings suggest that the gap between cascade and E2E system performance in both online and offline settings is narrowing. However, for some language pairs, the gap is still wide and therefore, additional work is warranted.  Our goal in the present ST survey is to offer valuable insight into this topic and drive advancements in ST research. We believe that such reviews will be interesting to researchers.





\bibliographystyle{elsarticle-harv} 
\bibliography{Reference}



\end{document}